\pdfoutput=1

\documentclass[11pt]{article}

\usepackage[final]{acl}

\usepackage{times}
\usepackage{latexsym}

\usepackage[T1]{fontenc}

\usepackage[utf8]{inputenc}

\usepackage{microtype}

\usepackage{inconsolata}

\usepackage{graphicx}

\usepackage{hyperref}
\usepackage{fancybox}
\usepackage{multirow}   
\usepackage{booktabs}
\usepackage{tabularx}
\usepackage{xcolor}
\usepackage{amsmath}
\usepackage{colortbl}
\usepackage{subcaption}
\usepackage{pifont}
\newcommand{\circnum}[1]{\raisebox{0pt}{\textcircled{\raisebox{-1.1pt} {#1}}}}

\newcommand{\method}{{\scshape Apa}}
\newcommand{\infogain}{{\scshape Infogain}}

\newcommand{\astfootnote}[1]{
    \let\oldthefootnote=\thefootnote
    \setcounter{footnote}{1}
    \renewcommand{\thefootnote}{\fnsymbol{footnote}}
    \footnotetext{#1}
    \let\thefootnote=\oldthefootnote
}

\title{Aligning Language Models to Explicitly Handle Ambiguity}

\author{
    Hyuhng Joon Kim\textsuperscript{\rm 1},
    Youna Kim\textsuperscript{\rm 1},
    Cheonbok Park\textsuperscript{\rm 2 3},
    Junyeob Kim\textsuperscript{\rm 1},\\
    \textbf{
    Choonghyun Park\textsuperscript{\rm 1},
    Kang Min Yoo\textsuperscript{\rm 1 2 4},
    Sang-goo Lee\textsuperscript{\rm 1 5},
    Taeuk Kim\textsuperscript{\rm 6 *}}\\
    \textsuperscript{\rm 1}Seoul National University,
    \textsuperscript{\rm 2}NAVER Cloud,
    \textsuperscript{\rm 3}KAIST AI, \\
    \textsuperscript{\rm 4}NAVER AI LAB,
    \textsuperscript{\rm 5}IntelliSys, Korea,
    \textsuperscript{\rm 6}Hanyang University\\
    \{heyjoonkim, anna9812, juny116, pch330, sglee\}@europa.snu.ac.kr\\
    \{cbok.park, kangmin.yoo\}@navercorp.com \\
    kimtaeuk@hanyang.ac.kr
}

\begin{document}

\maketitle
\begin{abstract}

    In interactions between users and language model agents, 
    user utterances frequently exhibit ellipsis (omission of words or phrases) or imprecision (lack of exactness) to prioritize efficiency.
    This can lead to varying interpretations of the same input based on different assumptions or background knowledge.
    It is thus crucial for agents to adeptly handle the inherent ambiguity in queries to ensure reliability.
    However, even state-of-the-art large language models (LLMs) still face challenges in such scenarios, primarily due to the following hurdles:
    (1) LLMs are not explicitly trained to deal with ambiguous utterances;
    (2) the degree of ambiguity perceived by the LLMs may vary depending on the possessed knowledge.
    To address these issues, we propose \textbf{Alignment with Perceived Ambiguity (\method{})}, a novel pipeline that aligns LLMs to manage ambiguous queries by leveraging their own assessment of ambiguity (i.e., \textbf{perceived ambiguity}).
    Experimental results on question-answering datasets demonstrate that 
    \method{} empowers LLMs to explicitly detect and manage ambiguous queries while retaining the ability to answer clear questions.
    Furthermore, our finding proves that \method{} excels beyond training with gold-standard labels, especially in out-of-distribution scenarios.
    The data and code are available at \url{https://github.com/heyjoonkim/APA}.

\end{abstract}
\astfootnote{Corresponding author.}

\section{Introduction}
\label{sec:introduction}

Large Language Models (LLMs) \citep{ouyang2022training,team2023gemini,achiam2023gpt} have demonstrated remarkable capabilities in text generation, 
proving particularly effective for question-answering (QA) tasks \citep{zhang2023survey,etezadi2023state}. 
QA systems in the wild frequently encounter unexpected user input, such as unanswerable \citep{kim-etal-2023-qa,yin-etal-2023-large} 
or ambiguous questions \citep{cole-etal-2023-selectively,lee-etal-2023-asking,kim-etal-2023-tree}.
To build an agent that is both reliable and user-friendly, it is essential for the model to robustly handle such inputs.
In this work, we seek to extend the scope of research to manage invalid inputs effectively.
Specifically, we focus on managing ``ambiguity'' \citep{gleason1963linguistics,mackay1967search}, 
which poses a significant challenge in Natural Language Processing (NLP) \citep{jurafsky1996probabilistic}.




\begin{figure}[t]
    \begin{center}
        \includegraphics[width=0.97\columnwidth]{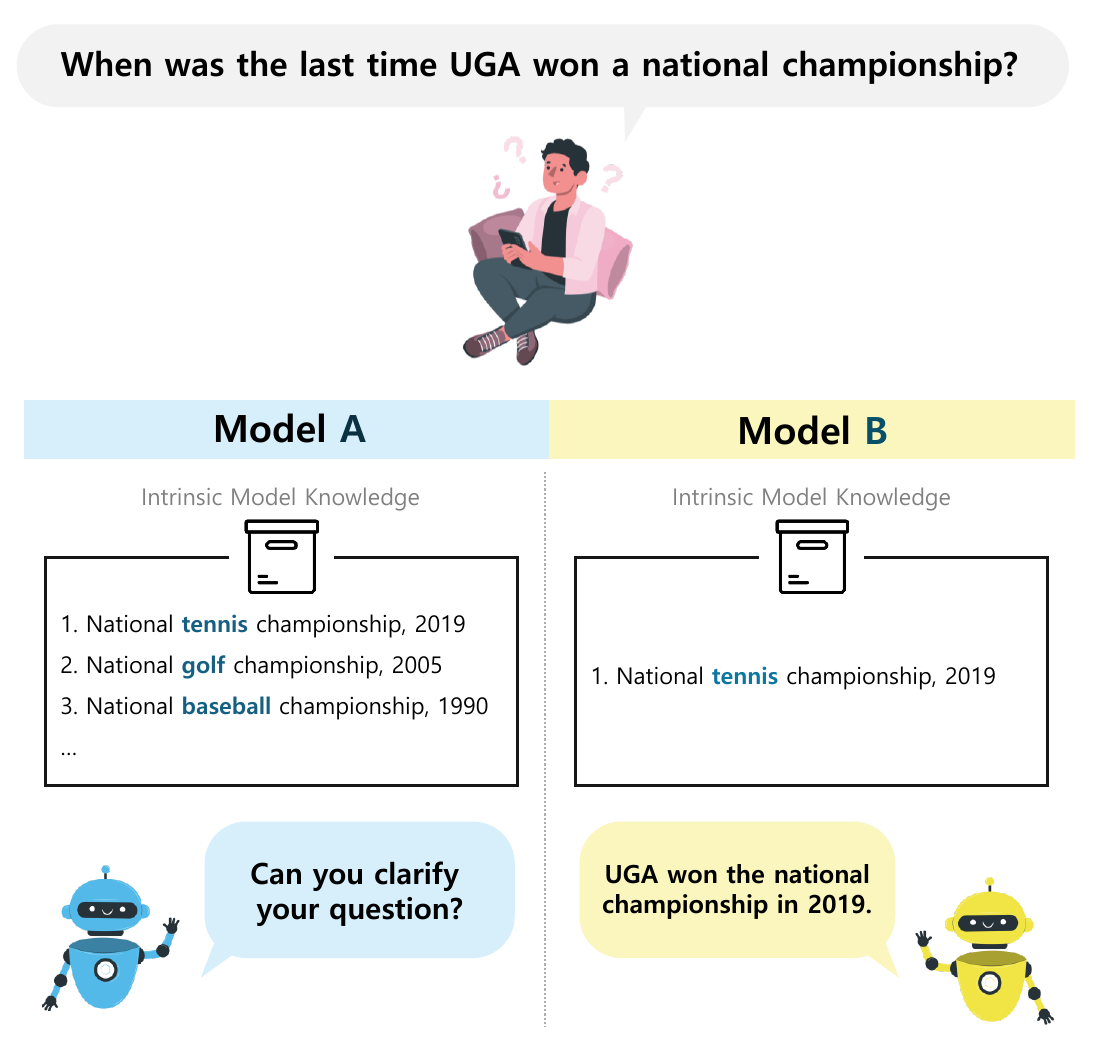}
          \caption{
            An example of an ambiguous query from AmbigQA.
            The term ``national championship'' poses diverse denotations, causing ambiguity.
            (Left) A model with diverse relevant knowledge
            might perceive the case as ambiguous.
            (Right) In contrast, 
            the query can be deemed unambiguous when the model lacks substantial related knowledge.
            Thus, the perceived ambiguity may differ depending on the model's intrinsic knowledge.
          }
          \label{fig:intro_image}
    \end{center}
\end{figure}

\textbf{Ambiguity} refers to cases where an expression conveys multiple denotations \citep{wasow2005puzzle}.
Users may pose queries with clear intentions that, 
possibly due to insufficient domain knowledge or omission during the utterance, result in ambiguous requests. 
If a model arbitrarily responds to such ambiguity, there is a risk of misinterpreting the user's original intent, potentially harming the model's reliability.
This is particularly evident in domains requiring high reliability, such as legal \citep{schane2002ambiguity,choi2024measuring} or medical \citep{stevenson2010disambiguation,gyori2022gilda}, 
where misinterpretations may lead to severe consequences.
Despite such importance, approaches to manage ambiguity robustly are still significantly unexplored.

  
Properly processing ambiguous inputs is challenging primarily due to the following two hurdles.
Firstly, models are \textbf{not trained to express ambiguity explicitly}.
Even if a model is capable of recognizing ambiguity,
confirming this recognition requires explicit cues from the model itself, such as expressing uncertainty or offering multiple interpretations.
The second challenge is that the \textbf{degree of ambiguity perceived by the model can vary} based on its intrinsic knowledge.
Consider the scenario depicted in Figure \ref{fig:intro_image}. 
The initial query is ambiguous
as the phrase ``national championship'' poses various denotations, 
such as ``national \textit{tennis} championship'' or ``national \textit{golf} championship''.
With comprehensive knowledge across possible denotations, a model can likely recognize the query's ambiguity (Figure \ref{fig:intro_image}, left). 
However, limited knowledge would lead the model to perceive the query as unambiguous (Figure \ref{fig:intro_image}, right).
Therefore, how a model interprets ambiguity hinges on its knowledge scope, which we define as \textbf{perceived ambiguity}.

To overcome these issues, this paper proposes \textbf{Alignment with Perceived Ambiguity (\method{})}---
a novel alignment pipeline for models to \textbf{explicitly handle} ambiguous queries by leveraging their \textbf{perceived ambiguity}.
Specifically, we design a proxy task that guides the model in utilizing its intrinsic knowledge for self-disambiguation of a given query.
We then quantify the information gained from this disambiguation as an implicit measure of the extent to which the model perceives the input as ambiguous.
This measure serves as a cue for ambiguous sample selection.
For the selected ambiguous query and its disambiguation, the model generates a clarification request regarding the ambiguity.
Finally, the model is trained to request explicit clarification in response to ambiguous queries.

Experimental results from a range of QA datasets demonstrate that \method{} enables a language model to properly handle ambiguous inputs
while maintaining its inherent capabilities of answering unambiguous queries.
Furthermore, we present three new datasets to provide a comprehensive framework for assessing ambiguity: AmbigTriviaQA, AmbigWebQuestions, and AmbigFreebaseQA.
These datasets facilitate a more extensive evaluation of models' robustness in addressing ambiguity, 
thus contributing to the further expansion of related research.

Our contributions can be summarized as follows:
\begin{enumerate}
    \item We propose a novel approach, Alignment with Perceived Ambiguity (\method{}), which enables language models to explicitly handle ambiguous inputs by leveraging perceived ambiguity.
    \item We introduce three new datasets--- AmbigTriviaQA, AmbigWebQuestions, and AmbigFreebaseQA---specifically designed to evaluate the model's capability of addressing ambiguity.
    \item Through empirical validation on multiple question-answering datasets, 
    we demonstrate that \method{} enables models to effectively handle ambiguous queries.
\end{enumerate}

\begin{figure*}[t]
    \centering
    \includegraphics[width=0.98\textwidth]{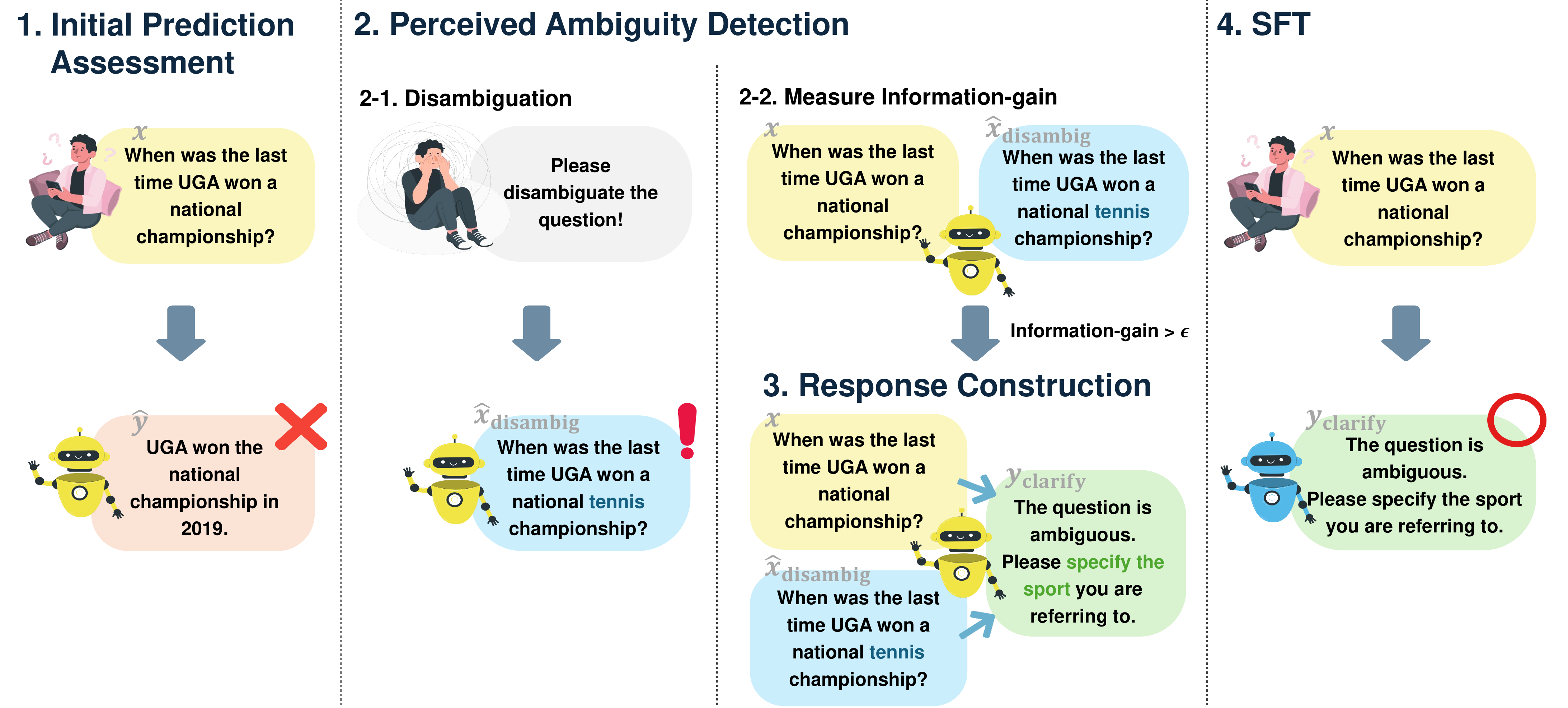} 

    \caption{
        The overall process of \method{}.
        We first select incorrect samples that the model currently fails to handle (Stage 1).
        The model then self-disambiguates these samples by leveraging its intrinsic knowledge.
        We measure the information gain (\infogain{}) between the initial input and the disambiguation, identifying  samples with high \infogain{} as ambiguous (Stage 2).
        Finally, the model generates a clarification request regarding the ambiguity (Stage 3),
        which is used as the label for training (Stage 4). 
    }
    \label{figure:main_figure}
\end{figure*}

\begin{figure}[t]
    \begin{center}
        \includegraphics[width=0.90\columnwidth]{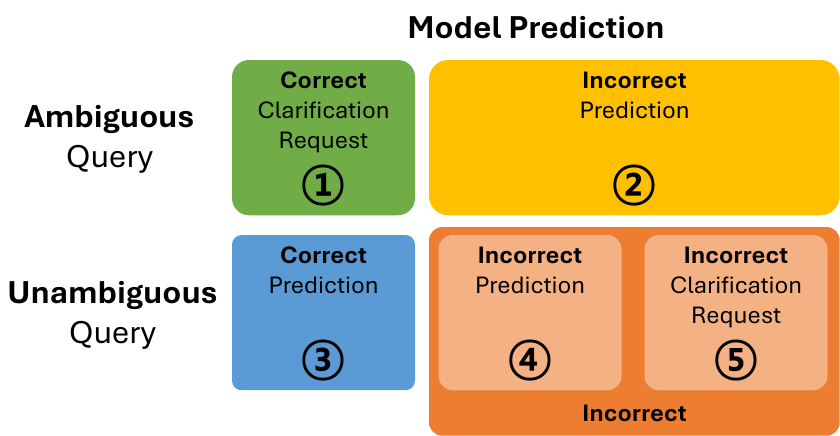}
          \caption{
            Illustration of five possible results from our scenario.  
            For ambiguous queries, the prediction is correct (\circnum{1}) if the model generates a clarification request;
            otherwise, all the other responses are classified as incorrect (\circnum{2}).
            When evaluating unambiguous queries, we compare the predictions to the ground-truth labels
            and categorize them as the correct prediction (\circnum{3}), incorrect prediction (\circnum{4}), or incorrect clarification request (\circnum{5}).
          }
          \label{fig:evaluation_metric}
    \end{center}
\end{figure}

\section{Related Work}
\label{sec:related_work}

\paragraph{Ambiguity in NLP}
An expression is ambiguous if it has two or more distinct denotations \citep{wasow2005puzzle}.
Ambiguity poses a significant challenge to NLP applications by obscuring the intended meaning of expressions, 
preventing models from accurately performing specific tasks.
Efforts to address this issue span across various domains, including  
machine translation \citep{pilault-etal-2023-interactive}, coreference resolution \citep{poesio-artstein-2005-reliability,yuan-etal-2023-ambicoref}, and natural language inference \citep{liu-etal-2023-afraid}.
The challenge intensifies in the scope of QA, 
as ambiguous questions may yield multiple answers that may not align with the user's initial intent. 
\citet{min-etal-2020-ambigqa} introduce the AmbigQA dataset to tackle ambiguity in open-domain QA
and \citet{stelmakh-etal-2022-asqa} expand it to long-form generation.
Furthermore, \citet{cole-etal-2023-selectively} demonstrate that quantifying sampling repetition presents a reliable uncertainty measure for ambiguity,
while \citet{kim-etal-2023-tree} generate tree-of-clarification (ToC) that refines input ambiguity.
While we share the goal of handling ambiguity, we propose a method of directly aligning the model.


\paragraph{Alignment of LLMs}

LLMs are typically trained through causal language modeling, 
a process essential for understanding and generating text of high fluency and consistency. 
To better harness these models, approaches have been developed to align them with human preferences \citep{DBLP:journals/corr/abs-1811-07871,ji2023ai}.
This has taken various forms, such as Reinforcement Learning from Human Feedback (RLHF) \citep{ouyang2022training, chakraborty2024maxminrlhf}, 
and Supervised Fine-tuning (SFT) \citep{dong2023raft, yang2023alignment, zhou2024lima}.
Previous works focused on preferences such as
helpfulness \citep{ding-etal-2023-enhancing, NEURIPS2023_949f0f8f, xu2024wizardlm},
safety \citep{bai2022constitutional, ji2023beavertails, liu2024enhancing},
and factuality \citep{yang2023alignment, tian2024finetuning}.
Building on this foundation, our research expands the scope
by focusing on aligning models to understand and manage ambiguity effectively.
\paragraph{Data Quality Control for Alignment}


Data-centric AI \citep{10.1145/2882903.2912574,10109290,kumar2024opportunities} highlights the importance of data quality in model training.
In the context of instruction-following techniques, LIMA \citep{zhou2024lima} 
demonstrates that effective model alignment can be achieved with just 1,000 high-quality, human-curated samples.
Similarly, AlpaGasus \citep{chen2024alpagasus} 
leverages only a small subset of the Alpaca dataset \citep{alpaca}, filtered by ChatGPT, for an effective alignment.
Various approaches for data selection have been explored, including those based on factors such as length and complexity \citep{liu2024what}, 
and gradient similarity from validation sets \citep{xia2024less}.
This work proposes a new viewpoint on data quality estimation: assessing how well data aligns models for ambiguity management.
For this purpose, we utilize the model's perceived ambiguity as an implicit cue for measuring data quality.




\section{Methodology}
\label{sec:methodology}

The primary goal of our research is to align models in a way that they can explicitly handle potentially ambiguous inputs, leveraging the model's perceived ambiguity. 
To this end, we propose \textbf{Alignment with Perceived Ambiguity (\method{})}, a four-stage alignment pipeline, illustrated in Figure \ref{figure:main_figure}.
In this section, we first formulate the problem and describe each stage in detail regarding the five possible results depicted in Figure \ref{fig:evaluation_metric}.
Further implementation details are stipulated in Appendix \ref{sec:implementation_details}.

\paragraph{Problem Formulation}

In this study, we focus on open-domain QA. 
The model $M$ is expected to generate a prediction $\hat{y}_{\text{unambig}}$ for an unambiguous query $x_{\text{unambig}}$ given a pre-defined inference template $t(\cdot)$.
$\hat{y}_{\text{unambig}}$ is compared to the ground-truth label $y$ and categorized as correct prediction (\circnum{3}), incorrect prediction (\circnum{4}), or incorrect clarification request (\circnum{5}).
As we expand our input scope to ambiguous queries\footnote{
    Separating ambiguous from unambiguous queries is inherently challenging due to subjective factors such as various perspectives and underlying assumptions.
    Despite the complexity, we simplify the problem and follow the pre-defined ambiguity from the training dataset for the alignment.
},
the model prediction for the ambiguous query $\hat{y}_{\text{ambig}}$ is anticipated to serve as a clarification request $y_{\text{clarify}}$
to resolve the ambiguity. 
This approach is grounded on the assumption that the user is best positioned to clarify their intent.\footnote{
    We explored alternatives for ambiguity management but found them to be impractical.
    For instance, arbitrarily selecting one of the valid answers may not accurately capture the user's intent. 
    Presenting all possible answers is often unfeasible due to the potentially vast number of valid responses.
}
$\hat{y}_{\text{ambig}}$ is considered correct (\circnum{1}) if it is a proper clarification request.
Otherwise, responses that fail to address the ambiguity are classified as incorrect (\circnum{2}).
The final objective of the alignment is to increase the number of samples corresponding to \circnum{1} 
while simultaneously maintaining or improving the proportion of responses classified as \circnum{3}.


\subsection{Initial Prediction Assessment}
\label{pipeline:explicit}
The initial stage focuses on identifying samples that the model currently fails to handle.
To do so, we compare the model's prediction with the ground-truth label, where samples are categorized based on accuracy.
Specifically, we assess the correctness by matching $\hat{y}_{\text{unambig}}$ with $y$ and $\hat{y}_{\text{ambig}}$ with $y_{\text{clarify}}$.
A total of $n$ correct samples, included in \circnum{1} and \circnum{3}, are collected as $D_{\text{correct}}=\{(x^i_{\text{correct}},y^i_{\text{correct}})\}^n_{i=1}$.
Incorrect samples falling under categories \circnum{2}, \circnum{4}, and \circnum{5} are unified as a separate dataset, $D_{\text{incorrect}}$.


\subsection{Perceived Ambiguity Detection}

This stage aims to identify samples from $D_{\text{incorrect}}$ that the model perceives as ambiguous. 
Given that it is challenging for the model to express ambiguity explicitly,
we construct a proxy task to estimate the ambiguity from the model's perspective.
Specifically, the model is prompted to self-disambiguate the given query $x$ and generate a disambiguation $\hat{x}_{\text{disambig}}$.
The model leverages its intrinsic knowledge related to $x$ to generate further details in this process.
If $x$ is underspecified and the model possesses related knowledge necessary to compensate,
then $\hat{x}_{\text{disambig}}$ would yield a higher certainty from the model's perspective.
On the other hand, if $x$ requires no specification or the model lacks the necessary knowledge, 
$\hat{x}_{\text{disambig}}$ would exhibit a similar level of uncertainty as $x$.
To quantify the uncertainty associated with $x$ and $\hat{x}_{\text{disambig}}$, we employ the model's average entropy \citep{malinin2021uncertainty,ABDAR2021243}. 
Formally, the entropy of an output distribution is defined as follows:
\begin{equation}
    \mathcal{H}_{x,i} = -\sum_{v \in \mathcal{V}} p_{x,i}(v)\log p_{x,i}(v)
\end{equation}
where $p_{x,i}(v)$ is the probability of the $i^{\text{th}}$ token $v$ of a sentence $x$ from the full vocabulary set $\mathcal{V}$.
The average entropy for $x$ can be defined as:
\begin{equation}
  \mathcal{H}_{x} = \frac{1}{N} \sum_{i} \mathcal{H}_{x,i}
\end{equation}
with $x$ composed of $N$-tokens.
We quantify the additional information gained from $\hat{x}_{\text{disambig}}$
by the difference in average entropy,
which we define as \textbf{information gain ({\scshape Infogain})}.
\begin{equation}
  \text{{\scshape Infogain}}_{x, \hat{x}_{\text{disambig}}} = \mathcal{H}_{x} - \mathcal{H}_{\hat{x}_{\text{disambig}}}
\end{equation}
A meaningful specification from $\hat{x}_{\text{disambig}}$ would result in a substantial \infogain{},
suggesting that the model perceives $x$ as ambiguous.
Regardless of the ground-truth ambiguity, samples 
with \infogain{} greater than the threshold $\epsilon$ are classified as ambiguous, denoted as $x_{\text{ambig}}$.


\subsection{Response Construction}
In this stage, we define $y_{\text{clarify}}$, which represents the clarification request the model should generate in response to an ambiguous query.
We explore two approaches for response generation: Fixed response and Generated response.

\paragraph{Fixed Response}
We utilize a pre-defined clarification request as $y_{\text{clarify}}$ for $x_{\text{ambig}}$.
Specifically, a list of clarification requests is pre-defined, and a single response is randomly selected as $y_{\text{clarify}}$ for each instance.

\paragraph{Generated Response}
The model is prompted to generate a clarification request specifying the source of the ambiguity.
To do so, we provide the model with $x_{\text{ambig}}$ and $\hat{x}_{\text{disambig}}$ to identify the aspect that causes the ambiguity,
thereby generating $y_{\text{clarify}}$ specific to the identified factor.

\subsection{Supervised Fine-Tuning (SFT)}

The objective of this stage is to construct datasets for the alignment.
Specifically, we label $m$ samples identified as ambiguous and construct an ambiguous dataset $D_{\text{ambig}}=\{(x^j_{\text{ambig}},y^j_{\text{clarify}})\}^m_{j=1}$,
where $y_{\text{clarify}}$ serves as the ground-truth label.
To prevent the potential loss of the model's existing knowledge, 
we also incorporate $D_{\text{correct}}$ for training.
The number of samples from both datasets are balanced so that $n=m$. 
The final training dataset is thus established as $D = D_{\text{correct}} + D_{\text{ambig}}$.
Utilizing the dataset $D=\{(x^k,y^k)\}^{n+m}_{k=1}$, the model is trained to generate 
$y$ for $x_{\text{unambig}}$ and $y_{\text{clarify}}$ for $x_{\text{ambig}}$,
employing the identical inference template $t(\cdot)$.
The model $M$ with parameter $\theta$ is trained as follows:
\begin{equation}
  \min_{\theta} \sum_{(x,y) \in D} \sum_{i=1}^{|y|} - \log M_{\theta}(y_i|y_{<i}, t(x))
\end{equation}
Two versions of \method{} are trained based on the type of $y_{\text{clarify}}$:
{\scshape Apa$_{\text{Fixed}}$} and {\scshape Apa$_{\text{Gen}}$},
which utilizes fixed and generated responses, respectively.

\section{Experimental Setting}
\label{sec:experimental_setting}

\subsection{Datasets}
The capability of the model to perform within the trained domain is pivotal.
However, for real-world applicability, the model must generalize to out-of-distribution (OOD) queries,
as queries that diverge from the training data are frequently confronted in practice.
Therefore, we utilize AmbigQA \citep{min-etal-2020-ambigqa} as the in-domain dataset for training and validation.
The dataset includes both ambiguous and unambiguous queries, with unambiguous queries labeled with ground-truth answers.
SituatedQA \citep{zhang-choi-2021-situatedqa} is used as a held-out OOD test dataset
with two different splits, denoted as SituatedQA-Geo and SituatedQA-Temp, 
each focusing on geographical and temporal ambiguities.
To further evaluate ambiguity across diverse QA domains, 
we have constructed three additional datasets: \textbf{AmbigTriviaQA}, \textbf{AmbigWebQuestions}, and \textbf{AmbigFreebaseQA},
each derived from TriviaQA \citep{joshi-etal-2017-triviaqa}, WebQuestions \citep{berant-etal-2013-semantic}, and FreebaseQA \citep{jiang-etal-2019-freebaseqa} respectively.
We prompt \texttt{gpt-4o}\footnote{\url{https://openai.com/index/hello-gpt-4o/}} to ambiguate the initial query from the original dataset and verify the generation.
To mitigate the potential biases in the validation process,
we further evaluate the verified samples with human annotators and select samples for the final dataset.
More details on the datasets and the construction process are described in Appendix \ref{sec:datasets_details}.

\subsection{Baselines}
\label{sec:baselines}
To evaluate the effectiveness of our approach, we introduce two sets of baselines: 
inference-only methods and trained methods. 
Specific implementation details are described in Appendix \ref{sec:baseline_details}.

\paragraph{Inference-Only Methods} 
\label{sec:inference_only}
Inference-only methods address ambiguity by utilizing different prompting strategies.
We employ direct prompting (\textbf{{\scshape Direct}}) as a fundamental baseline, 
applying a simple QA prompt. 
Furthermore, we explore ambiguity-aware prompting (\textbf{{\scshape Ambig-aware}}), 
which incorporates additional instructions on handling ambiguous inputs.
We also examine Sample Repetition (\textbf{{\scshape Sample Rep}}) \citep{cole-etal-2023-selectively} 
by measuring the consistency of the sampled generations.
Finally, we compare \textbf{{\scshape Self-Ask}} \citep{amayuelas2023knowledge}, 
where the model generates an answer
and subsequently determines the ambiguity based on the generation.

\paragraph{Trained Methods}
\label{sec:trained_methods}
Given the lack of directly comparable prior work,
we compare \method{} with fine-tuned baselines wherein the model is trained with the in-domain training set.
We follow the ambiguity as defined within the in-domain dataset, and train the model accordingly.
We compare \textbf{{\scshape Full-set}}, which applies the entire training dataset.
Furthermore, we compare two variations that leverages the equal number of training samples with \method{}.
\textbf{{\scshape Subset$_\text{Rand}$}} is trained on a randomly selected subset  with an equal number of ambiguous and unambiguous samples.
\textbf{{\scshape Subset$_\text{Ent}$}} applies the entropy of the model's prediction of the ambiguous query as the uncertainty measure.
Ambiguous samples with the most significant entropy are selected, and unambiguous samples are selected at random.

\begin{table*}[t]
    \centering    
    \resizebox{0.95\textwidth}{!}{

\begin{tabular}{lccccccccccc}
\toprule
Method & \begin{tabular}[c]{@{}c@{}}\# Training\\ Samples\end{tabular} & \multicolumn{2}{c}{\begin{tabular}[c]{@{}c@{}}SituatedQA-\\ Geo\end{tabular}} & \multicolumn{2}{c}{\begin{tabular}[c]{@{}c@{}}SituatedQA-\\ Temp\end{tabular}} & \multicolumn{2}{c}{\begin{tabular}[c]{@{}c@{}}Ambig-\\ TriviaQA\end{tabular}} & \multicolumn{2}{c}{\begin{tabular}[c]{@{}c@{}}Ambig-\\ WebQuestions\end{tabular}} & \multicolumn{2}{c}{\begin{tabular}[c]{@{}c@{}}Ambig-\\ FreebaseQA\end{tabular}} \\
\cmidrule(l){3-4} \cmidrule(l){5-6} \cmidrule(l){7-8} \cmidrule(l){9-10} \cmidrule(l){11-12}
 & & F1$_{u}$ & F1$_{a}$ & F1$_{u}$ & F1$_{a}$ & F1$_{u}$ & F1$_{a}$ & F1$_{u}$ & F1$_{a}$ & F1$_{u}$ & F1$_{a}$ \\ 
 \midrule
\multicolumn{12}{l}{\textbf{{\scshape Llama2 7B}}} \\ 
\midrule
{\scshape Direct} & 0 & 30.44 & 0.00 & 28.38 & 0.00 & 47.68 & 0.00 & 24.87 & 0.00 & 50.07 & 0.00 \\
{\scshape Ambig-aware} & 0 & 7.33 & 32.44 & 3.23 & 35.53 & 27.23 & 68.14 & 14.53 & 62.40 & 51.27 & 76.62 \\
{\scshape Sample Rep} & 0 & 6.83 & 34.43 & 8.28 & 38.43 & 53.11 & 72.63 & 13.31 & 69.21 & 63.11 & 78.70 \\
{\scshape Self-Ask} & 0 & 29.66 & 8.18 & 26.97 & 18.48 & 48.04 & 4.99 & 20.81 & 3.02 & 48.54 & 5.03 \\
{\scshape Subset$_{\text{Rand}}$} & 3,088 & 31.90 & 37.17 & 29.48 & 33.68 & 54.71 & 70.97 & 38.69 & 73.84 & 63.59 & 77.70 \\
{\scshape Subset$_{\text{Ent}}$} & 3,088 & 39.33 & 40.84 & \underline{34.28} & 34.62 & 58.83 & 74.98 & 42.39 & 75.86 & 72.18 & 83.89 \\
{\scshape Full-set} & 10,036 & 37.67 & 41.45 & 29.59 & 36.92 & 58.10 & 71.25 & 40.46 & 73.84 & 69.97 & 80.34 \\ 
\midrule
{\scshape Apa$_{\text{Fixed}}$} & 3,088 & \underline{39.99} & \underline{41.86} & 31.74 & \underline{39.63} & \textbf{62.97} & \underline{75.50} & \textbf{49.15} & \textbf{77.07} & \textbf{73.37} & \underline{84.19} \\
{\scshape Apa$_{\text{Gen}}$} & 3,088 & \textbf{41.01} & \textbf{43.10} & \textbf{34.38} & \textbf{41.89} & \underline{59.27} & \textbf{75.74} & \underline{47.26} & \underline{76.64} & \underline{73.18} & \textbf{84.90} \\ 
\midrule
\multicolumn{12}{l}{\textbf{{\scshape Mistral 7B}}} \\ 
\midrule
{\scshape Direct} & 0 & 11.29 & 0.00 & 15.34 & 0.00 & 33.19 & 0.00 & 17.85 & 0.00 & 31.37 & 0.00 \\
{\scshape Ambig-aware} & 0 & 3.66 & 26.01 & 8.43 & 22.48 & 26.26 & 48.43 & 8.39 & 30.52 & 32.96 & 54.91 \\
{\scshape Sample Rep} & 0 & 7.64 & 25.31 & 7.83 & 21.13 & 29.52 & 17.04 & 8.99 & 12.10 & 27.25 & 16.31 \\
{\scshape Self-Ask} & 0 & 11.29 & 0.00 & 15.34 & 0.00 & 33.19 & 0.00 & 17.85 & 0.00 & 31.37 & 0.00 \\
{\scshape Subset$_{\text{Rand}}$} & 1,382 & \underline{41.42} & 33.95 & 34.14 & 37.01 & 60.57 & 67.82 & 45.16 & 71.74 & 70.60 & 75.93 \\
{\scshape Subset$_{\text{Ent}}$} & 1,382 & \textbf{47.34} & 29.49 & 42.00 & 32.04 & 62.17 & 67.16 & 50.93 & 71.11 & 72.94 & 77.17 \\
{\scshape Full-set} & 10,036 & 35.99 & 41.28 & 31.16 & 33.72 & 66.67 & 76.38 & 41.83 & 74.72 & 76.98 & 84.67 \\ 
\midrule
{\scshape Apa$_{\text{Fixed}}$} & 1,382 & 38.43 & \underline{41.84} & \textbf{45.01} & \textbf{43.95} & \textbf{70.70} & \textbf{83.48} & \textbf{54.02} & \textbf{81.07} & \textbf{80.84} & \textbf{90.12} \\
{\scshape Apa$_{\text{Gen}}$} & 1,382 & 39.55 & \textbf{42.07} & \underline{43.29} & \underline{40.70} & \underline{67.73} & \underline{82.14} & \underline{51.41} & \underline{79.54} & \underline{80.27} & \underline{89.22} \\ 
\midrule
\multicolumn{12}{l}{\textbf{{\scshape Llama2 13B}}} \\ 
\midrule
{\scshape Direct}      & 0 & 30.44 & 0.00 & 29.69 & 0.00 & 46.43 & 0.00 & 27.59 & 0.00 & 49.17 & 0.00 \\
{\scshape Ambig-aware} & 0 & 5.99 & 33.10 & 4.22 & 36.66 & 24.80 & 68.19 & 4.81 & 65.28 & 43.81 & 73.40 \\
{\scshape Sample Rep}  & 0 & 11.57 & 32.85 & 16.56 & 37.87 & 49.93 & 72.44 & 7.89 & 67.26 & 61.05 & 79.33 \\
{\scshape Self-Ask}    & 0 & 30.44 & 0.00 & 29.69 & 0.00 & 46.43 & 0.00 & 27.59 & 0.00 & 49.17 & 0.00 \\
{\scshape Subset$_{\text{Rand}}$} & 3,216 & 33.11 & 36.87 & 28.57 & 37.84 & 63.19 & 73.52 & 44.31 & 72.99 & 70.40 & 78.29 \\
{\scshape Subset$_{\text{Ent}}$} & 3,216 & \textbf{40.19} & 38.39 & 31.03 & 38.00 & 64.95 & 76.03 & 48.70 & 77.43 & 73.38 & 81.93 \\
{\scshape Full-set} & 10,036 & \underline{37.58} & 38.39 & 29.41 & 34.37 & 68.33 & 76.82 & 47.20 & 75.27 & 76.56 & 83.00 \\ 
\midrule
{\scshape Apa$_{\text{Fixed}}$} & 3,216 & 31.31 & \textbf{40.23} & \textbf{36.45} & \textbf{42.18} & \textbf{70.83} & \textbf{80.99} & \textbf{53.69} & \textbf{79.22} & \textbf{79.92} & \textbf{88.03} \\
{\scshape Apa$_{\text{Gen}}$} & 3,216 & 34.04 & \underline{39.89} & \underline{31.72} & \underline{39.36} & \underline{69.25} & \underline{79.57} & \underline{52.96} & \underline{78.46} & \underline{79.80} & \underline{87.61} \\ 
\bottomrule
\end{tabular}

    }
    \caption{
        Experimental results for five different datasets. 
        We report the unambiguous and ambiguous F1-scores as F1$_{u}$ and F1$_{a}$, respectively.
        For each dataset, the \textbf{best method} is highlighted in bold and the \underline{second-best method} is underlined.
        \method{} outperforms all the baselines by utilizing the perceived ambiguity.
    }
    \label{table:main_results}
\end{table*}

\subsection{Evaluation Metrics}

A successful alignment should preserve the model's capability to handle unambiguous inputs while effectively managing ambiguous queries.
Based on the five possible results illustrated in Figure \ref{fig:evaluation_metric},
we define two distinct metrics to quantify such capabilities.
Further details of the evaluation process are described in Appendix \ref{sec:evaluation_details}.

\paragraph{Unambiguous Prediction F1 (F1$_{u}$)} 
The model must generate accurate answers to unambiguous queries while minimizing arbitrary responses to ambiguous queries. 
To measure this, we utilize the unambiguous prediction F1 score, 
which is the harmonic mean of precision ({\footnotesize$\frac{\circnum{3}}{\circnum{2}+\circnum{3}+\circnum{4}}$}) 
and recall ({\footnotesize$\frac{\circnum{3}}{\circnum{3}+\circnum{4}+\circnum{5}}$}) for ambiguous queries.

\paragraph{Ambiguity Detection F1 (F1$_{a}$)} 
Given an ambiguous input, the model should be able to detect them and generate clarification requests accordingly.
However, models may exhibit biased predictions toward clarification requests.
Taking these aspects into account, 
we evaluate the model's ambiguity detection capability with the F1-score, 
which captures both the precision ({\footnotesize$\frac{\circnum{1}}{\circnum{1}+\circnum{5}}$}) 
and recall({\footnotesize$\frac{\circnum{1}}{\circnum{1}+\circnum{2}}$}).

\subsection{Implementation Details}

For our experiments, we utilize {\scshape Llama2 7B $\&$ 13B} \citep{touvron2023llama}, and {\scshape Mistral 7B} \citep{jiang2023mistral}.
We employ QLoRA \citep{dettmers2023qlora} to facilitate efficient training.
Results are averaged over three different random seeds.


\begin{figure}[t]
    \includegraphics[width=\linewidth]{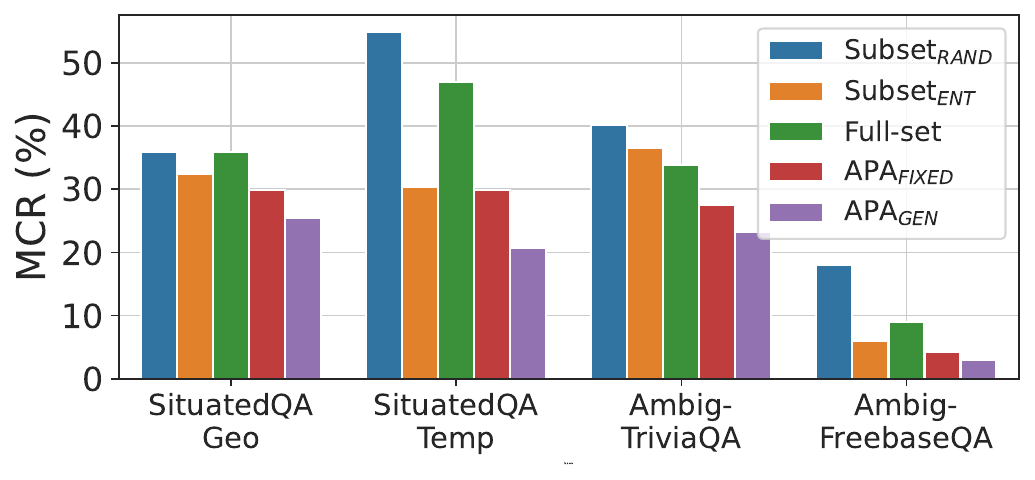} 
    \caption{
       Misaligned Clarification Request Rate (MCR) of trained methods. 
       Low MCR indicates that the model 
       retains its intrinsic knowledge 
       even after the alignment process. 
       In all instances, \method{} exhibits the lowest MCR.
    }
    \label{figure:misclassification}
\end{figure}



\begin{figure}[t]
    \centering
    \begin{subfigure}{0.49\columnwidth}
        \includegraphics[width=\columnwidth]{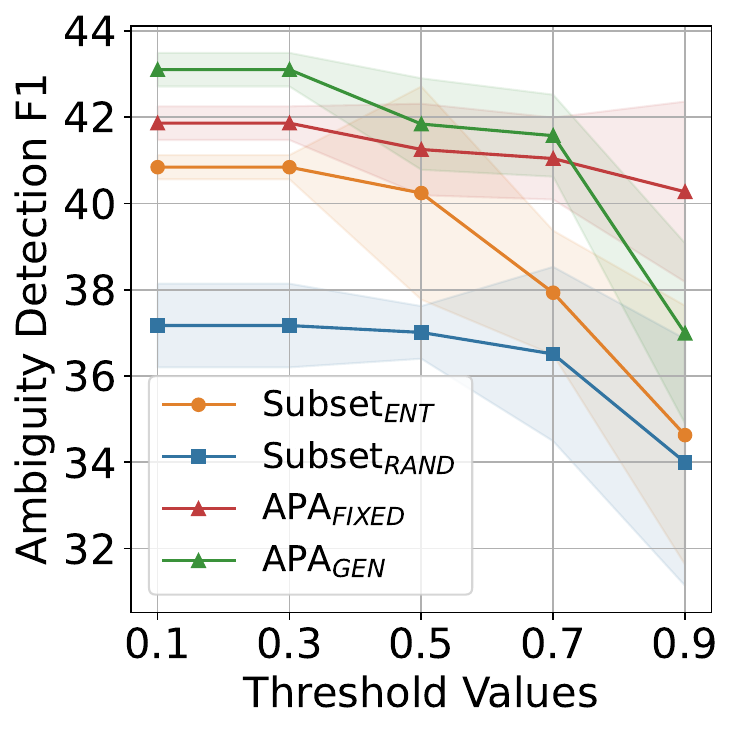}
        \caption{SituatedQA-Geo}
        \label{figure:threshold_ablation_situatedqa_geo}
    \end{subfigure}
    \hfill
    \begin{subfigure}{0.49\columnwidth}
        \includegraphics[width=\columnwidth]{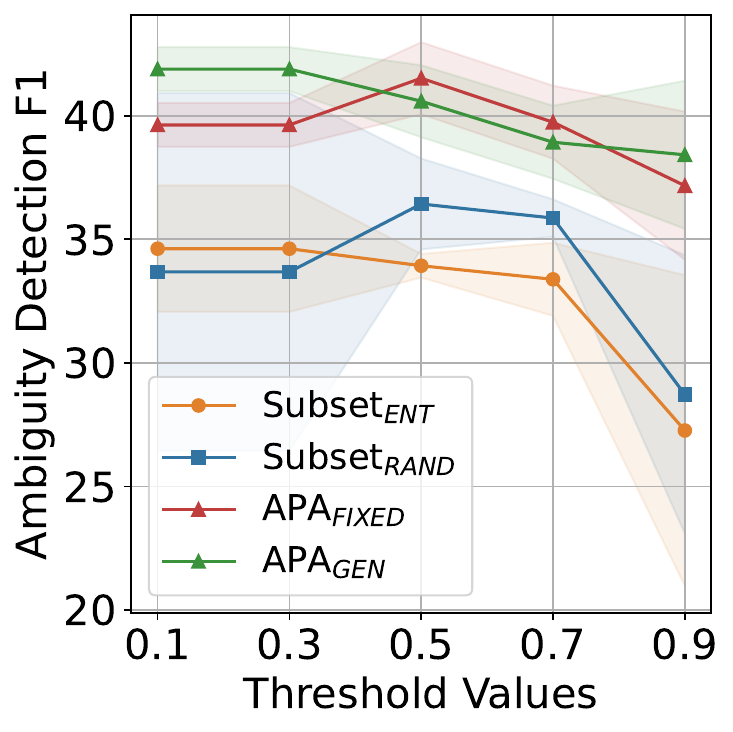}
        \caption{SituatedQA-Temp}
        \label{figure:threshold_ablation_situatedqa_temp}
    \end{subfigure}
    \hfill
    \begin{subfigure}{0.49\columnwidth}
        \includegraphics[width=\columnwidth]{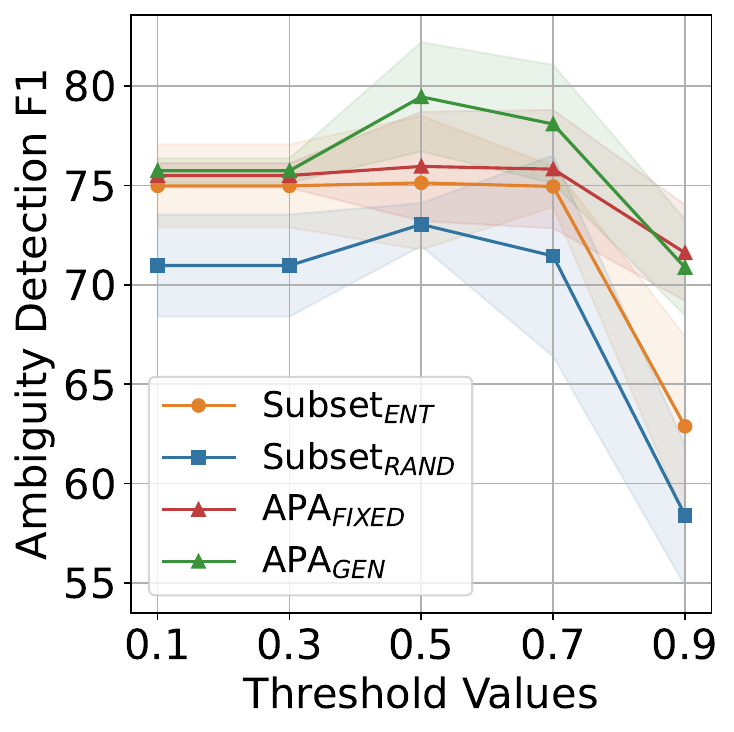}
        \caption{AmbigTriviaQA}
        \label{figure:threshold_ablation_ambig_triviaqa}
    \end{subfigure}
    \hfill
    \begin{subfigure}{0.49\columnwidth}
        \includegraphics[width=\columnwidth]{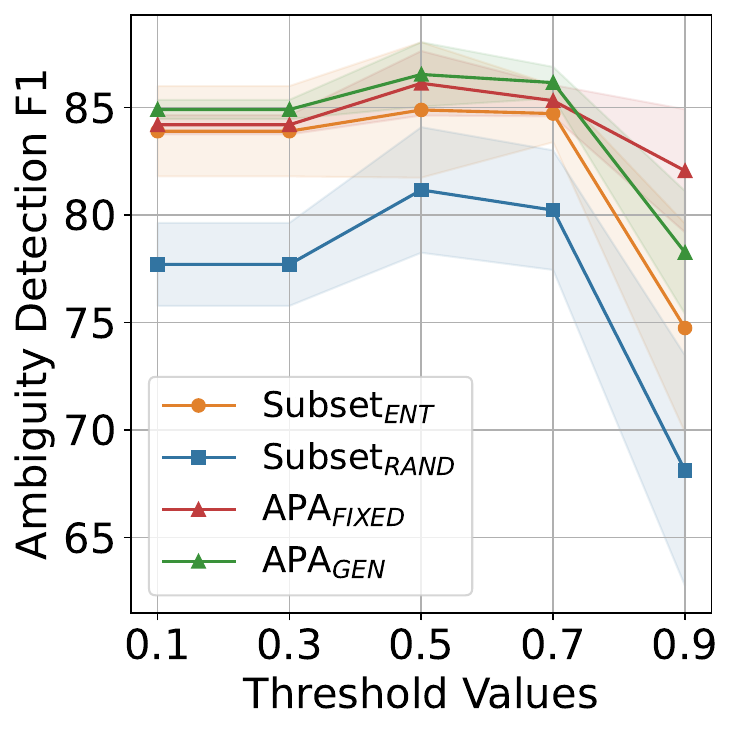}
        \caption{AmbigFreebaseQA}
        \label{figure:threshold_ablation_ambig_freebaseqa}
    \end{subfigure}

    \caption{
        Changes in the F1$_{a}$ score according to the threshold value.
        Regardless of the threshold value, 
        {\scshape Apa} consistently outperforms all the baselines.
    }
    \label{figure:threshold_ablation}
\end{figure}

\section{Experimental Results}
\label{sec:experimental_results}

The main results are presented in Table \ref{table:main_results}. 

\textbf{Inference-only methods exhibit significant limitations in handling ambiguous queries}.
{\scshape Direct} fails to manage ambiguous queries, as evidenced by its consistent zero F1$_{a}$ scores.
{\scshape Ambig-aware} and {\scshape Sample Rep} demonstrate a strong bias towards clarification requests, exhibiting deficient F1$_{u}$.
{\scshape Self-Ask} displays a subpar F1$_{a}$,
indicating it is challenging to resolve ambiguity by just ``asking'' the model without task-specific training.

\textbf{Trained methods present enhanced performance compared to inference-only approaches.} 
Specifically, {\scshape Subset$_{\text{Rand}}$} exhibits improved performance across both metrics compared to inference-only methods.
{\scshape Full-set} demonstrates superior performance among the baselines, leveraging the entire training set.
Notably, {\scshape Subset$_{\text{Ent}}$} surpasses {\scshape Subset$_{\text{Rand}}$} by a large margin
and even outperforms {\scshape Full-set} in some datasets.
The results of {\scshape Subset$_{\text{Ent}}$} verify that entropy is capable of capturing ambiguity to some extent and is beneficial when incorporated into the alignment process.
\textbf{\method{} achieves superior performance across all datasets.}
Despite employing an identical inference template, \method{} achieves a notable enhancement in F1$_{u}$ compared to {\scshape Direct}.
This improvement is especially surprising considering that \method{} was trained on $D_{\text{correct}}$, which consists of samples that the model is already capable of handling.
Moreover, \method{} consistently outperforms across all the datasets in terms of F1$_{a}$, achieving gains up to 6 points.
The results highlight the effectiveness of leveraging perceived ambiguity for alignment, enhancing generalization and robustness.
When compared to {\scshape Subset$_{\text{Ent}}$}, the improvement of \method{} suggests that \infogain{} provides better quantification of ambiguity than entropy.
The efficacy of leveraging only the data perceived ambiguous, comprising approximately 32\% in the {\scshape Llama2} family and 13\% in {\scshape Mistral}, 
again emphasizes the importance of data quality over quantity \citep{zhou2024lima,chen2024alpagasus}.
Furthermore, {\scshape Apa$_{\text{Fixed}}$} generally exhibits enhanced performance compared to {\scshape Apa$_{\text{Gen}}$}. 
This is because {\scshape Apa$_{\text{Gen}}$} engages in a more challenging task of generating specific clarification requests.


\definecolor{first}{HTML}{f8bc04}
\definecolor{second}{HTML}{fdd768}
\definecolor{third}{HTML}{fcf2cf}

\begin{table}[t]
    \centering    
    \resizebox{\linewidth}{!}{

        \begin{tabular}{lcccc}
            \toprule
            
            Method & 
            \begin{tabular}[c]{@{}c@{}}SituatedQA-\\Geo\end{tabular} & 
            \begin{tabular}[c]{@{}c@{}}SituatedQA-\\Temp\end{tabular} & 
            \begin{tabular}[c]{@{}c@{}}Ambig-\\TriviaQA\end{tabular} & 
            \begin{tabular}[c]{@{}c@{}}Ambig-\\FreebaseQA\end{tabular} \\
            
            \midrule
            
            {\scshape Rand} & 
            \cellcolor{third}\begin{tabular}[c]{@{}c@{}}39.31 \\ \footnotesize{(1.28)} \end{tabular} & 
            \cellcolor{third}\begin{tabular}[c]{@{}c@{}}38.34 \\ \footnotesize{(0.44)} \end{tabular} &
            \cellcolor{third}\begin{tabular}[c]{@{}c@{}}72.05 \\ \footnotesize{(0.58)} \end{tabular} & 
            \cellcolor{third}\begin{tabular}[c]{@{}c@{}}81.28 \\ \footnotesize{(1.88)} \end{tabular} \\
            
            {\scshape Min} & 
            \begin{tabular}[c]{@{}c@{}}34.95 \\ \footnotesize{(1.71)} \end{tabular} & 
            \begin{tabular}[c]{@{}c@{}}36.03 \\ \footnotesize{(0.90)} \end{tabular} &
            \begin{tabular}[c]{@{}c@{}}70.30 \\ \footnotesize{(1.50)} \end{tabular} & 
            \begin{tabular}[c]{@{}c@{}}79.19 \\ \footnotesize{(2.02)} \end{tabular} \\
            
            {\scshape Max} & 
            \cellcolor{second}\begin{tabular}[c]{@{}c@{}}40.96 \\ \footnotesize{(0.71)} \end{tabular} & 
            \cellcolor{second}\begin{tabular}[c]{@{}c@{}}39.33 \\ \footnotesize{(0.88)} \end{tabular} &
            \cellcolor{second}\begin{tabular}[c]{@{}c@{}}73.95 \\ \footnotesize{(1.03)} \end{tabular} & 
            \cellcolor{second}\begin{tabular}[c]{@{}c@{}}82.23 \\ \footnotesize{(1.31)} \end{tabular} \\
            
            {\scshape Apa}& 
            \cellcolor{first}\begin{tabular}[c]{@{}c@{}}43.10 \\ \footnotesize{(0.39)} \end{tabular} & 
            \cellcolor{first}\begin{tabular}[c]{@{}c@{}}41.89 \\ \footnotesize{(2.02)} \end{tabular} &
            \cellcolor{first}\begin{tabular}[c]{@{}c@{}}75.74 \\ \footnotesize{(1.52)} \end{tabular} & 
            \cellcolor{first}\begin{tabular}[c]{@{}c@{}}84.90 \\ \footnotesize{(0.40)} \end{tabular} \\
            
            \bottomrule
        \end{tabular}

    }
    \caption{
        Average and standard deviation (in parentheses) of F1$_{a}$ scores of different data selection methods.
        The \colorbox{first}{first}, \colorbox{second}{second}, and \colorbox{third}{third} best results are highlighted.
        Results show that utilizing \infogain{} regardless of the ground-truth ambiguity
        is effective for data selection.
    }
    \label{table:implicit_measure}
\end{table}

\begin{table*}[ht]
    \centering    
    \resizebox{0.98 \textwidth}{!}{

    {\footnotesize
    \begin{tabular}{cl}
    
    \toprule
    Type & \multicolumn{1}{c}{Generations}  \\ 
    \midrule 
    
    $x$ & 
    How many pages in a brave new world? \\
    $\hat{x}_{\text{disambig}}$ & 
    How many pages in \textbf{the 1932 edition of the book} brave new world \textbf{by Aldous Huxley}? \\ 
    $y_{\text{clarify}}$ & 
    Your question is ambiguous. \underline{Which edition} of the book are you interested in? \\ 
    \midrule
    $x$ & 
    Who was the commander of the british forces in boston? \\
    $\hat{x}_{\text{disambig}}$ & 
    Who was the commander of the british forces in boston \textbf{during the american revolution?} \\ 
    $y_{\text{clarify}}$ & 
    Your question seems ambiguous. Can you be more \underline{specific about the event or time?} \\      
    \bottomrule
    
    \end{tabular}
    }
}

    \caption{
        Examples of generated $y_{\text{clarify}}$ and $\hat{x}_{\text{disambig}}$ from the initial query $x$.
        \textbf{Additional specification} from the disambiguation is highlighted in bold and the \underline{specification of the clarification requests} are underlined.
    }
    \label{table:case_study}
\end{table*}

\section{Ablation Study}
\label{sec:ablations}
In this section, we perform a series of ablation studies to further evaluate \method{}.
Unless otherwise specified, all experiments are conducted on {\scshape Llama2 7B} across four datasets: SituatedQA-Geo, SituatedQA-Temp, AmbigTriviaQA, and  AmbigFreebaseQA.
Additional details are stipulated in Appendix \ref{sec:ablation_details}.

\subsection{Analysis on Sample-level Misalignment}
\newcommand{\misalignment}{MCR}
The alignment process of generating clarification requests for ambiguous queries may lead to a potential trade-off, 
where the model incorrectly generates clarification requests for unambiguous inputs that were previously well-handled.
To assess such a case, we define \textbf{Misaligned Clarification Request Rate (\misalignment{})}, 
which measures the proportion of unambiguous samples 
that were correctly answered (\circnum{3} in Figure \ref{fig:evaluation_metric}) before training but incorrectly shifted to erroneously generating clarification requests (\circnum{5} in Figure \ref{fig:evaluation_metric}) after alignment.
A low \misalignment{} is desirable, representing that the model preserves its existing capabilities even after the alignment.
We can observe from Figure \ref{figure:misclassification} that, overall, \method{} consistently demonstrates the lowest \misalignment{}, indicating that
the model successfully learns to handle ambiguity while effectively preserving the existing capabilities.

\subsection{The Effect of Threshold Values}

The number of training samples used for alignment depends on the threshold value $\epsilon$.
To understand the impact of $\epsilon$ on performance, we conduct an analysis by applying different $\epsilon$ for ambiguous data selection.
We compare {\scshape Subset$_{\text{Ent}}$} and {\scshape Subset$_{\text{Rand}}$}, 
each with an equal number of training samples.
Figure \ref{figure:threshold_ablation} presents the F1$_{a}$ scores
measured under different $\epsilon$.
In general, larger $\epsilon$ reduces the data available for training, 
resulting in declined performance.
{\scshape Subset$_{\text{Rand}}$} consistently demonstrates subpar performance, 
whereas {\scshape Subset$_{\text{Ent}}$} is a strong baseline across all scenarios.
Nevertheless, \method{} outperforms all the baselines across different $\epsilon$ values.

\subsection{Impact of \infogain{} for Data Selection}
\label{sec:ablation_3}

For a deeper analysis of \infogain{} on data selection within \method{}, 
we conducted an ablation study by varying the criteria for selecting ambiguous data.
With the correct dataset $D_{\text{correct}}$ held constant,
we alter the strategies of selecting $m$ ambiguous samples as follows:

\begin{itemize}
    \item \textbf{Random Selection ({\scshape Rand})} 
    We randomly select $m$ ground-truth ambiguous samples.    
    \item \textbf{\infogain{}-based Selection}
    We explore two different selection methods leveraging \infogain{}:
    \textbf{{\scshape Max}} selects top-$m$ samples with the largest \infogain{} from the ground-truth ambiguous samples.
    \textbf{{\scshape Min}} selects the bottom-$m$ samples with the minimum \infogain{}
    among those that are ground-truth ambiguous.
\end{itemize}


\method{} differs from the baselines by utilizing samples perceived as ambiguous, 
allowing the potential inclusion of ground-truth unambiguous samples.

Table \ref{table:implicit_measure} demonstrates the overall results.
{\scshape Rand} consistently lags behind {\scshape Max} by a margin of 1 to 4 points.
The disparity underscores the effectiveness of data selection based on \infogain{}, 
even with ground-truth ambiguous samples.
Moreover, \method{} outperforms all the baselines across all the datasets.
Notably, even though the perceived ambiguity
does not always coincide with ground-truth ambiguity,
results show that exploiting model-perceived ambiguity
significantly enhances alignment.
{\scshape Min} demonstrates the worst performance
among the methods evaluated.
We speculate that this decline is because the training samples with low \infogain{} are perceived as unambiguous, yet are trained as ambiguous.
This misalignment likely accounts for the degradation in performance.




\subsection{Case Study}
\label{sec:case_study}

Table \ref{table:case_study} demonstrates examples of generated disambiguation $\hat{x}_{\text{disambig}}$ and the clarification request $y_{\text{clarify}}$ from the query $x$.
We can observe that the model generates factual specifications about the query leveraging its intrinsic knowledge (e.g., \textit{1932 edition of the book}).
Furthermore, given $x$ and $\hat{x}_{\text{disambig}}$, the model successfully generates a clarification request, specifically mentioning the factor that causes the ambiguity (e.g., \textit{Which edition}).
Further examples of disambiguations and failure cases are in Appendix \ref{sec:additional_case_study}.



\section{Conclusion}
\label{sec:conclusion}

In this work, we present a novel alignment pipeline, dubbed \textbf{Alignment with Perceived Ambiguity (\method{})}, designed to enhance the ability of LLMs to 
address ambiguities within queries, 
leveraging the model's intrinsic knowledge. 
Our method employs an implicit measure \infogain{} to quantify the ambiguity perceived by the model itself.
The model learns to effectively manage (un)ambiguous queries through alignment based on this metric.
Experimental results demonstrate the effectiveness of \method{}, which outperforms all the baselines across various QA datasets.
As a future avenue, we plan to explore extending this methodology to broader domains and more complex types of ambiguities, further solidifying the role of LLMs in managing the inherent uncertainty present in NLP tasks.

\section*{Limitations}
The scope of our research is mainly focused on short-form QA tasks.
The research scope could be expanded to long-form generation tasks such as detailed reasoning.
Furthermore, there are cases when a query becomes ambiguous by considering additional contexts, e.g., cases in conversational QA \citep{guo2021abgcoqa}.
As our research focuses solely on situations where a single query is given, future work may consider scenarios where additional context is provided to the model.
For experiments, we explore the most widely used models for evaluation, 
specifically {\scshape Llama2} and {\scshape Mistral}.
Despite this, a more comprehensive evaluation encompassing a broader range of LLMs could have enriched our findings, providing insights across different architectures and capabilities.
Larger-scale models may exhibit different tendencies and, therefore, should be explored in future research.
Furthermore, our work mainly focuses on supervised fine-tuning (SFT) as the alignment method.
However, alternative methods, such as Reinforcement Learning from Human Preference (RLHF) \citep{ouyang2022training} or Direct Preference Optimization (DPO) \citep{rafailov2023direct},
could offer distinct advantages toward our objective.

\section*{Acknowledgement}

This work was partly supported by SNU-NAVER Hyperscale AI Center and 
Institute of Information \& communications Technology Planning \& Evaluation (IITP) grant funded by the Korea government (MSIT) [NO.RS-2021-II211343, Artificial Intelligence Graduate School Program (Seoul National University), 
No.RS-2020-II201373, Artificial Intelligence Graduate School Program (Hanyang University), 
NO.RS-2021-II212068, Artificial Intelligence Innovation Hub (Artificial Intelligence Institute, Seoul National University)]

\bibliography{anthology,custom}

\appendix


\section{Implementations Details}
\label{sec:implementation_details}

\subsection{Pipeline Details}
\label{sec:pipeline_details}
For initial prediction assessment (Stage 1), we utilize the same inference template as {\scshape Direct} (Table \ref{template:direct})
and disambiguate the given query with the template from Table \ref{template:disambiguation}.
We use the greedy generation for the disambiguation.
The threshold $\epsilon$ is empirically set to 0.1 for selecting ambiguous inputs.
When balancing training set size, if $n>m$, we randomly select $m$ samples from $D_{\text{correct}}$,
where $n = |D_{\text{correct}}|$ and $m = |D_{\text{ambig}}|$.
If $n<m$, we select $n$ samples from $D_{\text{ambig}}$ with the largest \infogain{}.
For {\scshape Apa$_{\text{Gen}}$}, we use the template from Table \ref{template:clarification} to generate specific clarification requests for each ambiguous queries.
Furthermore, for {\scshape Apa$_{\text{Fixed}}$}, we randomly set $y_{\text{clarify}}$ from the following pre-defined phrases : 
[\texttt{The questions is ambiguous.
Please clarify your question.
Your question is ambiguous.
Can you clarify your question?
Your question is not clear.
Can you clarify your question please?}]

\subsection{Training Details}
For training, we applied AdamW optimizer \citep{Loshchilov2019DecoupledWD} with a batch size of 32.
We selected the model with the best performance in the validation set from learning rates \{\texttt{1e-3, 5e-4, 1e-4}\} and training epochs \{\texttt{1, 2, 3}\}.
All the experiments were implemented with Pytorch \citep{paszke2019pytorch} and Huggingface Transformers library \citep{wolf-etal-2020-transformers}.
For efficient training, we applied QLoRA from Huggingface PEFT library \citep{peft} with {\itshape r=4} and {\itshape alpha=16}.
The training takes about half an hour on a single Tesla V100 GPU.
All experiments are averaged over three different random seeds. 
The full results of {\scshape Apa} and trained baseline methods with the standard deviation are demonstrated in Table \ref{table:full_results}.

\begin{table}[t]
    \centering    

   \begin{tabularx}{\linewidth}{X}
    \toprule
\ttfamily
Answer the following question. \\
\ttfamily
Question: \textcolor{brown}{<question>} \\
\ttfamily
Answer:
\\

    \bottomrule
\end{tabularx}

    \caption{
        Direct prompting template.
    }
    \label{template:direct}
\end{table}

\begin{table}[t]
    \centering    

   \begin{tabularx}{\linewidth}{X}
    \toprule

\ttfamily
Evaluate the clarity of the input question.
If the question is ambiguous, enhance it by adding specific details such as relevant locations, time periods, or additional context needed to resolve the ambiguity.
For clear questions, simply repeat the query as is.\\\\

\ttfamily
Example:\\
\ttfamily
Input Question: When did the Frozen ride open at Epcot?\\
\ttfamily
Disambiguation: When did the Frozen ride open at Epcot?\\\\
\ttfamily
Input Question: What is the legal age of marriage in the USA?\\
\ttfamily
Disambiguation: What is the legal age of marriage in each state of the USA, excluding exceptions for parental consent?\\\\

\ttfamily
Input Question: \textcolor{brown}{<question>} \\
\ttfamily
Disambiguation:\\

    \bottomrule
\end{tabularx}

    \caption{
        Disambiguation template used in Perceived Ambiguity Detection Stage of {\scshape Apa}.
        We provide 2-shot demonstrations from AmbigQA train set.
    }
    \label{template:disambiguation}
\end{table}

\begin{table}[t]
    \centering    

   \begin{tabularx}{\linewidth}{X}
    \toprule

\ttfamily
Engage with the provided ambiguous question by extracting the key point of ambiguity, and interactively ask for clarification based on the disambiguated question.\\\\

\ttfamily
Example 1:\\
\ttfamily
Ambiguous Question: Who won?\\
\ttfamily
Disambiguation: Who won the 2020 U.S. presidential election?\\
\ttfamily
Clarification Request: Your question seems ambiguous. Could you specify which competition or event you are asking about?\\\\

\ttfamily
Example 2:\\
\ttfamily
Ambiguous Question: What’s the weather like?\\
\ttfamily
Disambiguation: What’s the weather like in Miami today?\\
\ttfamily
Clarification Request: Your question is ambiguous. Where are you interested in the weather report for?\\\\

\ttfamily
Ambiguous Question: \textcolor{brown}{<ambiguous question>}\\
\ttfamily
Disambiguation: \textcolor{brown}{<disambiguation>}\\
\ttfamily
Clarification Request:\\

    \bottomrule
\end{tabularx}

    \caption{
        Template for generating clarification request for the given ambiguous query.
        The model is prompted to extract the factor that causes the ambiguity and generate a clarification request based on the extracted factor.
    }
    \label{template:clarification}
\end{table}

\begin{table}[t]
    \centering    
    \resizebox{\columnwidth}{!}{

\begin{tabular}{lcccc}

\toprule
Dataset & \multicolumn{2}{c}{\begin{tabular}[c]{@{}c@{}}Train \end{tabular}}& \multicolumn{2}{c}{\begin{tabular}[c]{@{}c@{}}Validation / Test\end{tabular}}\\
 & Unambig. & Ambig. & Unambig. & Ambig. \\
\midrule
AmbigQA & 5,287 & 4,749 & 830 & 1,172 \\
SituatedQA-Geo & - & - & 506 & 129 \\
SituatedQA-Temp & - & - & 2,795 & 876 \\
AmbigTriviaQA & - & - & 500 & 500 \\
AmbigWebQuestions & - & - & 500 & 500 \\
AmbigFreebaseQA & - & - & 500 & 500 \\
\bottomrule
\end{tabular}

    }
    \caption{
        Number of ambiguous and unambiguous samples for each datasets.
        We utilize AmbigQA for in-domain training and validation.
        The rest of the datasets are evaluated as OOD test sets.
    }
    \label{table:dataset_size}
\end{table}

\begin{table}[t]
    \centering    

   \begin{tabularx}{\linewidth}{X}
    \toprule
    
\ttfamily
Please make the following question ambiguous. Your task is to introduce ambiguity by altering the specificity of the noun phrase or omitting crucial details from the statement. Keep the rest of the sentence unchanged except for the modified sections. Generate only the revised statement. \\\\
\ttfamily
Question: \textcolor{brown}{<question>} \\
\ttfamily
Ambiguation:
\\

    \bottomrule
\end{tabularx}

    \caption{
        Template to ambiguate the input query for dataset construction.
        We prompt \texttt{gpt-4o} for the generation.
    }
    \label{template:ambig_triviaqa_generation}
\end{table}

\begin{table}[t]
    \centering    

   \begin{tabularx}{\linewidth}{X}
    \toprule
    
\ttfamily
An ambiguous question has multiple valid answers. Is the following question ambiguous with multiple possible answers? Answer only in Yes or No. \\\\
\ttfamily
Question: \textcolor{brown}{<ambiguous generation>} \\\\
\ttfamily
Yes or No:
\\
    \bottomrule
\end{tabularx}

    \caption{
        Template for validating the generated ambiguated queries.
        We prompt \texttt{gpt-4o} for the validation.
        Samples with the output "Yes" are considered a valid ambiguation.
    }
    \label{template:ambig_triviaqa_validation}
\end{table}

\begin{table}[t]
    \centering    

   \begin{tabularx}{\linewidth}{X}
    \toprule
    
\ttfamily
You are given an ambiguous question and its possible ambiguation. Please verify whether the ambiguous question poses proper ambiguity. An ambiguous question must have multiple valid answers. \\\\
\ttfamily
Original Question: \textcolor{brown}{<original question>} \\
\ttfamily
Ambiguous Question: \textcolor{brown}{<ambiguated question>} \\\\
\ttfamily
Yes or No:
\\
    \bottomrule
\end{tabularx}

    \caption{
        Instructions for human validation for dataset construction. 
        Samples selected as "Yes" are considered a valid ambiguation.
    }
    \label{template:human_evaluation}
\end{table}

\begin{table}[t]
    \centering    

   \begin{tabularx}{\linewidth}{X}
    \toprule
    \ttfamily
Answer the following question. If the question is ambiguous, it is proper to answer with ``The question is ambiguous''.\\
\ttfamily
Question: \textcolor{brown}{<question>} \\
\ttfamily
Answer:
\\

    \bottomrule
\end{tabularx}

    \caption{
        Ambiguity-aware prompting. 
        We explicitly describe how to handle ambiguity.
    }
    \label{template:ambiguity_aware}
\end{table}

\begin{table}[t]
    \centering    

   \begin{tabularx}{\linewidth}{X}
    \toprule
    
\ttfamily
Answer the following question. Given the question and answer, is the question ambiguous or unambiguous? Answer only ambiguous or unambiguous. \\
\ttfamily
Question: \textcolor{brown}{<question>} \\
\ttfamily
Answer: \textcolor{brown}{<generated answer>} \\\\
\ttfamily
Is the question ambiguous or unambiguous? Answer only ambiguous or unambiguous. \\
\ttfamily
Ambiguous or Unambiguous:
\\

    \bottomrule
\end{tabularx}

    \caption{
        Verification template for {\scshape Self-Ask}.
        With the generated answer and the original question, 
        the model is prompted to verify the ambiguity of the initial query.
    }
    \label{template:self_ask}
\end{table}



\section{Dataset Overview}
\label{sec:datasets_details}

\subsection{Dataset Details}
This section stipulates the details of the datasets we used in the experiments.
The statistics of ambiguous and unambiguous samples for each dataset is specified in Table \ref{table:dataset_size}.

\paragraph{AmbigQA} \citep{min-etal-2020-ambigqa}
is a derivative of the Natural Questions dataset \citep{kwiatkowski-etal-2019-natural}, designed to verify ambiguous data points. 
The dataset covers diverse sources of ambiguity, such as event and entity references.
The dataset consists of pre-defined ambiguous and unambiguous queries, where unambiguous queries are labeled with ground-truth answers.
We set AmbigQA as the in-domain dataset and utilize it for training and validation.
Specifically, we follow the ambiguity defined by the dataset and train the model
to generate ground-truth answers for unambiguous queries and pre-defined clarification requests for ambiguous queries. 
Further training details are stipulated in Appendix \ref{sec:baseline_details}.

\paragraph{SituatedQA} \citep{zhang-choi-2021-situatedqa}
focuses explicitly on temporal and geographic ambiguity from the input query.
As the cause of ambiguity and its construction process are distinct,
we assess performance on the temporal and geographic split separately,
denoted as Temp and Geo, respectively.

\paragraph{TriviaQA} \citep{joshi-etal-2017-triviaqa}
consists of question-answer-evidence triplets collected from Wikipedia and the web.
For our experiments, we only utilize the question-answer pairs.
We ambiguiate the subset of TriviaQA to build AmbigTriviaQA.

\paragraph{WebQuestions} \citep{berant-etal-2013-semantic}
is a question-answering dataset that uses Freebase as the knowledge base.
The dataset consists of questions from the Google Suggest API and then answers obtained from Amazon Mechanical Turk.
In creating AmbigWebQUestions, we applied ambiguity to the subset of WebQuestions.

\paragraph{FreebaseQA} \citep{jiang-etal-2019-freebaseqa}
is an open-domain QA over the Freebase knowledge graph.
The question-answer pairs are collected from various sources such as TriviaQA, QuizBalls, and QuizZone.
AmbigFreebaseQA is derived from the subset of FreebaseQA.


\subsection{Dataset Construction Details}
To further examine the model's capability to interpret and generate responses to intentionally ambiguous queries,
we constructed AmbigTriviaQA, AmbigWebQuestions, and AmbigFreebaseQA 
by ambiguating the TriviaQA, WebQuestions, and FreebaseQA, respectively.
We first prompt \texttt{gpt-4o} to ambiguate the original question with the template from Table \ref{template:ambig_triviaqa_generation}.
To further validate the generation and control the dataset's quality,
we again prompt \texttt{gpt-4o} for secondary verification.
We utilize the template in Table \ref{template:ambig_triviaqa_validation} and collect samples verified as ambiguous.
Validating the generations from the same model may pose unnecessary biases.
To mitigate the potential biases in the validation process, we evaluate the verified samples with human annotators and select samples for the final dataset. (Table \ref{template:human_evaluation})
This human-in-the-loop data construction ensures the quality and fairness of the dataset.
The process yielded 1,000 question-answer pairs, with 500 ambiguous and 500 unambiguous pairs.
Examples from AmbigTriviaQA are demonstrated in Table \ref{table:ambig_triviaqa_example}.



\section{Baseline Details}
\label{sec:baseline_details}
In this section, we describe implementation details of the baselines.

\paragraph{{\scshape Direct}}
We make a direct inference using the template from Table \ref{template:direct}.
The greedy generation result with temperature 0 is used for evaluation.

\paragraph{{\scshape Ambig-aware}}
We utilize the template from Table \ref{template:ambiguity_aware}, where we explicitly describe how to handle ambiguity.
Identically, we use the greedy generations for evaluation.

\paragraph{{\scshape Sample Rep}}
The template from Table \ref{template:direct} is used to generate a single greedy generation and ten sampled generations with sampling temperature of 1.0.
We quantify the rate of sampled generations that match the greedy generation as the uncertainty measure,
where 1.0 is the most certain and 0.0 being the least certain.
Samples with the measure below a specific threshold are considered ambiguous.
For instance, if three out of ten samples exactly match the greedy generation, then the uncertainty for the given query is 0.3.
We empirically select a threshold that demonstrates the best F1$_{u}$ and F1$_{a}$ with the least trade-off.

\paragraph{{\scshape Self-Ask}}
We initially prompt the model with the template from Table \ref{template:direct} and generate a greedy generation.
Then, the initial query and the generated answer are utilized with the template from Table \ref{template:self_ask}
and prompt the model to verify the query's ambiguity.
We modified the prompt from \citet{amayuelas2023knowledge} so that the model can specifically focus on ambiguity.
The ambiguity detection is determined based on the model's final verification of "Yes" or "No".


\paragraph{{\scshape Full-set}}
The entire training set is utilized for training. 
Following {\scshape Apa$_{\text{Fixed}}$}, we label the ground-truth ambiguous samples with pre-defined clarification requests as $y_{\text{clarify}}$.
(Pre-defined clarification requests are listed in Appendix \ref{sec:pipeline_details}.)
The model is trained to generate $y$ for $x_{\text{unambig}}$ and $y_{\text{clarify}}$ for $x_{\text{ambig}}$
with the inference template from Table \ref{template:direct}.
 
\paragraph{{\scshape Subset$_{\text{Rand}}$}}
The training method is identical to {\scshape Full-set}, but {\scshape Subset$_{\text{Rand}}$} utilizes a subset of the training set.
We randomly select $|D|$ samples from the training data, with the equal number ($|D|/2$) of ambiguous and unambiguous samples.

\paragraph{{\scshape Subset$_{\text{Ent}}$}}
The training of {\scshape Subset$_{\text{Rand}}$} is identical to {\scshape Subset$_{\text{Rand}}$} except the ambiguous sample selection method.
When $x_{\text{ambig}}$ is given, we measure the entropy of the generated result from the model.
A high entropy value indicates that the model is uncertain about the prediction of the ambiguous query.
Therefore, among the $x_{\text{ambig}}$ in the train set, 
we select $|D|/2$ samples with the highest output entropy and use them as ambiguous samples.


\section{Evaluation Details}
\label{sec:evaluation_details}
In this section, we describe the evaluation details of our experiments. 
We utilize the greedy generation from the model for the evaluation.

\subsection{Unambiguous Query Evaluation}
For unambiguous queries, we measure the quality of the generation by employing RougeL\footnote{\url{https://huggingface.co/spaces/evaluate-metric/rouge}} \citep{lin-och-2004-automatic} with all the possible valid answers.
The prediction from the model is regarded as correct if the score is above 0.3.

\subsection{Ambiguous Query Evaluation}
For ambiguous questions, we expect the model to generate clarification requests. 
Since there are various ways to express clarification requests, 
we use the following phrases to detect the requests.
The presence of pre-defined ambiguity-related phrases in the model's output is treated as a successful detection. 
The pre-defined phrases are the follows:
[\texttt{ambiguous, ambig, unclear, not clear, not sure, confused, confusing, vague, uncertain, doubtful, doubt, questionable, clarify, not clear}]


\section{Details of Ablation Experiments}
\label{sec:ablation_details}
\subsection{Details of Sample-level Misalignment Analysis}
To measure Misaligned Clarification Request rate (MCR), we start with a base model (e.g., {\scshape Llama2 7B} or {\scshape Mistral 7B}) which has not undergone any alignment training.
We prompt the model using the template in Table \ref{template:direct} and select the correct, unambiguous samples.
Subsequently, we evaluate the aligned models, such as {\scshape Full-set}, {\scshape Subset$_{\text{Ent}}$}, or {\scshape Apa$_{\text{Gen}}$}, leveraging these pre-selected samples.
We then count the cases where the aligned model's predictions shifted from providing correct answers to generating wrong clarification requests post-alignment.
MCR is measured as the proportion of these shifted samples relative to the total number of initially correct, unambiguous samples.
The metric quantified the extent to which the model's alignment process leads to unnecessary clarification requests for previous well-handled unambiguous queries.

\begin{table}[t]
    \centering    
    \resizebox{\columnwidth}{!}{

        \begin{tabular}{l|ccccc}
        
            \toprule
            Threshold & 0.1 & 0.3 & 0.5 & 0.7 & 0.9 \\
            \# Samples & 3,088 & 3,088 & 1,860 & 886 & 396 \\
            \bottomrule
        \end{tabular}

    }
    \caption{
        Number of training samples for different threshold values.
        We vary the threshold value from 0.1 to 0.9.
    }
    \label{table:samples_per_threshold}
\end{table}

\subsection{Details of Threshold Ablation}
To measure the performance with different threshold values, we apply $\epsilon \in$ \{\texttt{0.1, 0.3, 0.5, 0.7, 0.9}\}.
The number of selected samples for training is illustrated in Table \ref{table:samples_per_threshold}.

\begin{figure}[t]
    \includegraphics[width=\linewidth]{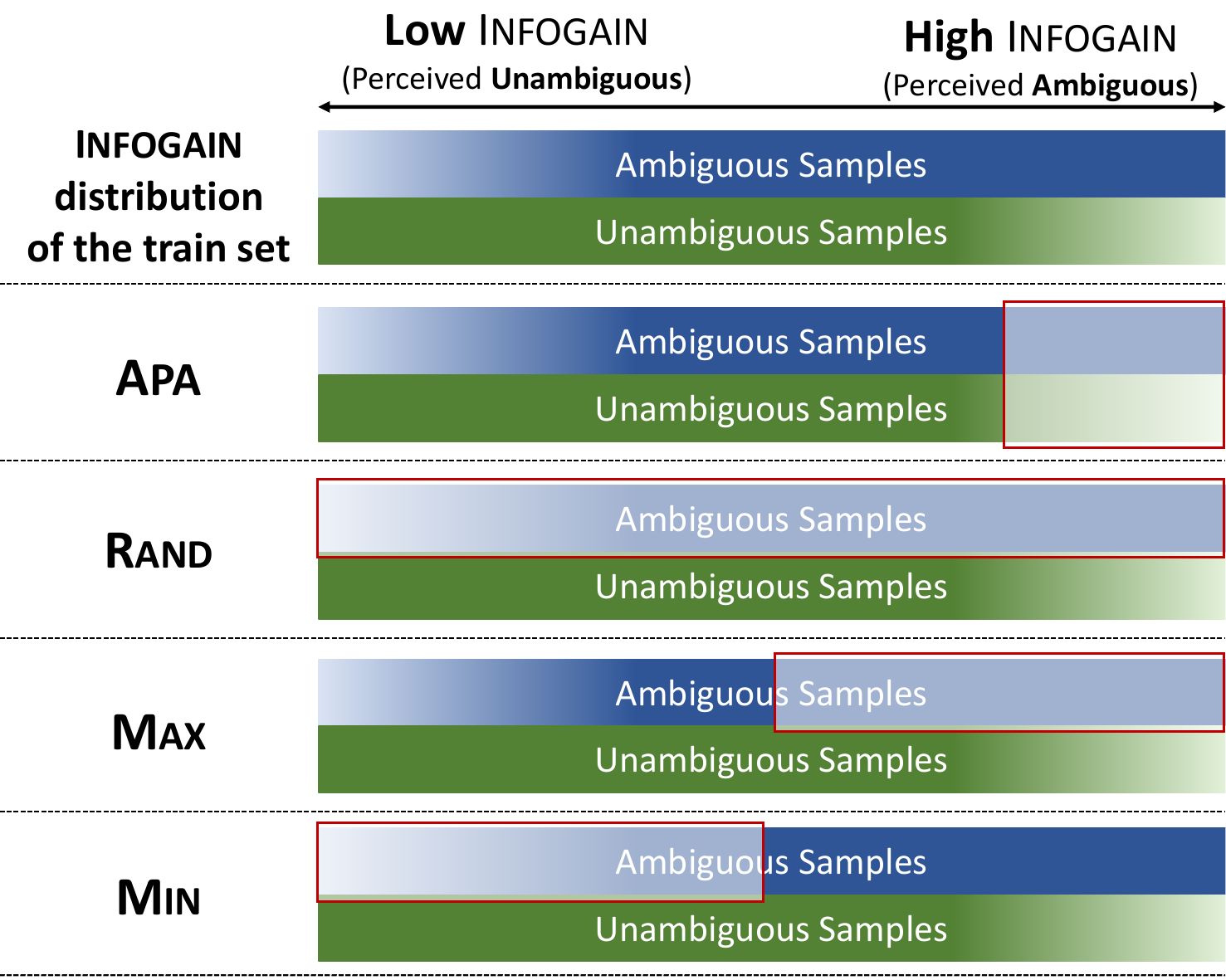} 
    \caption{
        Illustration of ground-truth ambiguous and unambiguous samples sorted by the \infogain{}.
        We highlight the chosen samples for each data selection method.
        \method{} selects samples with the largest \infogain{} regardless of the ground-truth ambiguity.
        On the other hand, baseline methods select training data from ground-truth ambiguous samples with different selection strategies.
    }
    \label{figure:ablation_3}
\end{figure}

\subsection{Details of Data Selection Ablation}
This section details the data selection methods from Section \ref{sec:ablation_3}, with the corresponding visualization in Figure \ref{figure:ablation_3}.
Consider the case where the ground-truth ambiguous and unambiguous queries are sorted based on their \infogain{}.
\method{} selects $m$-samples with the largest \infogain{} regardless of the ground-truth ambiguity, focusing on perceived ambiguity.
In contrast, {\scshape Rand} randomly selects $m$-samples as ambiguous from the ground-truth ambiguous queries (highlighted in blue in Figure \ref{figure:ablation_3}).
{\scshape Max} and {\scshape Min} select top-$m$ and bottom-$m$ samples regarding the \infogain{} from the ground-truth ambiguous queries, respectively.
Unlike the baseline methods, which only consider the ground-truth ambiguity, \method{} leverages the perceived ambiguity, which may not always align with the ground-truth ambiguity.


\section{Additional Case Studies}
\label{sec:additional_case_study}

\subsection{Failure Cases Before Alignment}
Table \ref{table:failure_cases} demonstrates generations by models before alignment for ambiguous queries from SituatedQA-Geo. 
Given the diverse denotations of the query, each model interprets the query differently based on their intrinsic knowledge.
For instance, the first question is ambiguous due to the numerous possible ``revolution'' it could reference.
Each model interprets ``revolution'' differently:
{\scshape Llama2 7B} as the ``\textit{Russian} revolution'', {\scshape Mistral 7B} as the ``\textit{French} revolution'', and {\scshape Llama2 13B} as the ``\textit{American} Revolutionary War''.
Consequently, each model generates factual responses corresponding to its interpretation.
We regard this phenomenon as problematic since the user likely has a specific ``revolution'' in mind while querying the model. 
However, the model may misinterpret the input and generate responses not aligned with the user's intended reference.
Consequently, this misalignment can lead to providing incorrect or irrelevant answers.

\subsection{Case Study of Disambiguations}
Table \ref{table:disambiguation_examples} demonstrates examples of initial query $x$ and its disambiguation $\hat{x}_{\text{disambig}}$.
The first example is when $x$ is inherently ambiguous, yet the model perceives it as unambiguous. 
Specifically, the model generates hallucination ("in the 1960s") where the song "don't mess around with jim" was originally released in 1972.
This non-factual generation would not provide any information gain to the model, classifying $x$ as ambiguous. 
In such a case, $x$ should be considered "unknown" with no related knowledge within the model.
The second and third examples are correctly classified,
as the model properly applies its intrinsic knowledge to perceive ambiguity.
Regardless of the quantity of additional context generated, 
the model is capable of verifying its ambiguity.
The last example is a misclassification as ambiguous.
Despite disambiguation provides factually correct information ("1932 novel" and "by Aldous Huxley") for "brave new world",
we speculate that the misclassification may arise from the existence of various media, such as movies and songs or even different versions of the book, sharing the title "brave new world".

\subsection{Failure Cases of Clarification Request Generation}
Table \ref{table:case_study_failure_cases} presents failure cases of clarification request generation.
Even when the model successfully provides valid disambiguation (e.g., \textit{in the usa} or \textit{in 2015}),
in some cases the model fails to consider the aspect that causes the ambiguity while generating clarification requests.
For example, the first case generates "What is the book the title refers to?", which does not address the relevant ambiguity.
Furthermore, the second example only requests for clarification and fails to provide further specifications regarding the ambiguity.

\begin{table*}[t]
    {\small
        \begin{tabularx}{\textwidth}{XX}
            \toprule
            \multicolumn{1}{c}{Original Question} & 
            \multicolumn{1}{c}{Ambiguated Question} \\ 
            \midrule
            Who wrote the 19th century \textbf{novel ‘Anna Karenina’}? & 
            Who wrote a 19th century \textbf{novel}? \\
            \midrule
            What was \textbf{President Gerald Ford's} middle name? & 
            What was the middle name of a \textbf{former U.S. president?} \\
            \midrule
            Where in England was \textbf{actor Nigel Hawthorne} born? & 
            Where in the UK was \textbf{the actor} born? \\ 
            \bottomrule
        \end{tabularx}
    }
    \caption{
            Examples of the original question and its ambiguation from AmbigTriviaQA.
            The \textbf{ambiguated phrase} is highlighted in bold. 
        }
    \label{table:ambig_triviaqa_example}
                
\end{table*}

\begin{table*}[t]
    \centering    
    \small
    \begin{tabularx}{0.98\textwidth}{XXXX}

\toprule
Question & Llama2 7B & Mistral 7B & Llama2 13B  \\ 
 \midrule
When did the revolution begin? & The revolution began in 1917. \textcolor{lightgray}{\textit{(Russian revolution)}} & The revolution began in 1789. \textcolor{lightgray}{\textit{(French revolution)}}& The revolution began in 1775. \textcolor{lightgray}{\textit{(American Revolutionary War)}} \\ 
\midrule
When did we host the world cup? & 1992 \textcolor{lightgray}{\textit{(Not factual)}} & 1994 \textcolor{lightgray}{\textit{(15$^{th}$ World Cup)}} & 1996. \textcolor{lightgray}{\textit{(Not factual)}} \\ 
 \midrule
Who was the first elected president? & George Washington
 & George Washington & The first elected president of India was Dr. Rajendra Prasad. \\ 
 \bottomrule

\end{tabularx}

    \caption{
        Model predictions of ambiguous queries from SituatedQA-Geo before alignment.
        Ambiguous query, due to its variety of denotations, leads the model to interpret the query differently based on its intrinsic knowledge,
        consequently generating diverse answers.
    }
    \label{table:failure_cases}
\end{table*}

\begin{table*}[t]
    \centering    
    \small
    \begin{tabularx}{0.98\textwidth}{cccX}

\toprule
\begin{tabular}[c]{@{}c@{}}Model\\ Prediction\end{tabular} & \begin{tabular}[c]{@{}c@{}}Ground\\ Truth\end{tabular} & Type & \multicolumn{1}{c}{Generated Text}  \\ 
\midrule
\multirow{2}{*}{Unambig.} & \multirow{2}{*}{Ambig.} & $x$ & Who sings don't mess around with jim? \\
 &  & $\hat{x}_{\text{disambig}}$ & Who sings don't mess around with jim, \textbf{in the 1960s?} \\ 
 \midrule
\multirow{3}{*}{Unambig.} & \multirow{3}{*}{Unambig.} & $x$ & Who is winner in bigg boss season 5 kannada? \\
 &  & \multirow{2}{*}{$\hat{x}_{\text{disambig}}$} & Who is the winner of the fifth season of the kannada \textbf{version of the indian reality television series bigg boss?} \\ 
 \midrule
\multirow{2}{*}{Ambig.} & \multirow{2}{*}{Ambig.} & $x$ & How many jury members in a criminal trial? \\
 &  & $\hat{x}_{\text{disambig}}$ & How many jury members are required in a criminal trial \textbf{in the united states?} \\ 
 \midrule
\multirow{2}{*}{Ambig.} & \multirow{2}{*}{Unambig.} & $x$ & How many pages in a brave new world? \\
 &  & $\hat{x}_{\text{disambig}}$ & How many pages are in the \textbf{1932 novel} brave new world \textbf{by aldous huxley?} \\ 
 \bottomrule

\end{tabularx}

    \caption{
        Example of initial query $x$ and its disambiguation $\hat{x}_{\text{disambig}}$.
        \textbf{Additional specification} from the model is in bold.
        Unambig. and Ambig. refers to Unambiguous and Ambiguous, respectively.
    }
    \label{table:disambiguation_examples}
\end{table*}

\begin{table*}[t]
    \centering    
    \resizebox{0.75 \textwidth}{!}{

        {\small
            \begin{tabular}{cl}
            
                \toprule
                Type & \multicolumn{1}{c}{Generations}  \\ 
                \midrule 
                $x$ & 
                Where was the old man and the sea published? \\
                $\hat{x}_{\text{disambig}}$ & 
                Where was the old man and the sea published \textbf{in the usa}? \\ 
                $y_{\text{clarify}}$ & 
                Your question is ambiguous. What is the book the title refers to? \\ 
                
                \midrule
                $x$ & 
                Who won last triple crown of horse racing? \\
                $\hat{x}_{\text{disambig}}$ & 
                Who won the triple crown of horse racing \textbf{in 2015}? \\ 
                $y_{\text{clarify}}$ & 
                I’m not sure about your question, could you  provide some more information. \\ 
                \bottomrule
            
            \end{tabular}
        }
    }

    \caption{
        Failure cases of generated clarification request $y_{\text{clarify}}$ from the initial query $x$ and its disambiguation $\hat{x}_{\text{disambig}}$.
        \textbf{Additional specification} from the disambiguation is highlighted in bold.
        Despite the correct disambiguations, the model fails to generate clarification requests regarding the ambiguity.
    }
    \label{table:case_study_failure_cases}
\end{table*}
\begin{table*}[t]
    \centering    
    \resizebox{0.95 \textwidth}{!}{
        
        \begin{tabular}{lcccccc}
        \toprule
        Method & \multicolumn{2}{c}{{\scshape Llama2 7B}} & \multicolumn{2}{c}{{\scshape Mistral 7B}} & \multicolumn{2}{c}{{\scshape Llama2 13B}} \\
         & F1$_{u}$ & F1$_{a}$ & F1$_{u}$ & F1$_{a}$ & F1$_{u}$ & F1$_{a}$ \\ 
         \midrule
        \multicolumn{7}{l}{\textbf{SituatedQA-Geo}} \\
        \midrule
        {\scshape Subset$_{\text{Rand}}$} &
        31.90 \footnotesize{(3.29)} & 
        37.17 \footnotesize{(0.97)} & 
        \underline{41.42} \footnotesize{(3.08)} & 
        33.95 \footnotesize{(1.62)} & 
        33.11 \footnotesize{(3.21)} & 
        36.87 \footnotesize{(0.85)} \\
        {\scshape Subset$_{\text{Ent}}$} & 
        39.33 \footnotesize{(3.77)} & 
        40.84 \footnotesize{(0.28)} & 
        \textbf{47.34} \footnotesize{(1.41)} & 
        29.49 \footnotesize{(4.36)} & 
        \textbf{40.19} \footnotesize{(0.95)} & 
        38.39 \footnotesize{(1.80)} \\
        {\scshape Full-set} & 
        37.67 \footnotesize{(1.87)} & 
        41.45 \footnotesize{(1.19)} & 
        35.99 \footnotesize{(1.18)} & 
        41.28 \footnotesize{(0.40)} & 
        \underline{37.58} \footnotesize{(1.71)} & 
        38.39 \footnotesize{(1.01)} \\ 
        \midrule
        {\scshape Apa$_{\text{Fixed}}$} & 
        \underline{39.99} \footnotesize{(0.96)} & 
        \underline{41.86} \footnotesize{(0.39)} & 
        38.43 \footnotesize{(1.17)} & 
        \underline{41.84} \footnotesize{(0.39)} & 
        31.31 \footnotesize{(3.32)} & 
        \textbf{40.23} \footnotesize{(0.40)} \\
        {\scshape Apa$_{\text{Gen}}$} & 
        \textbf{41.01} \footnotesize{(0.89)} & 
        \textbf{43.10} \footnotesize{(0.39)} & 
        39.55 \footnotesize{(5.14)} & 
        \textbf{42.07} \footnotesize{(1.13)} & 
        34.04 \footnotesize{(4.59)} & 
        \underline{39.89} \footnotesize{(2.10)} \\

        
         \midrule
        \multicolumn{7}{l}{\textbf{SituatedQA-Temp}} \\
        \midrule
        {\scshape Subset$_{\text{Rand}}$} & 
        29.48 \footnotesize{(7.72)} & 
        33.68 \footnotesize{(7.24)} & 
        34.14 \footnotesize{(5.02)} & 
        37.01 \footnotesize{(0.82)} & 
        28.57 \footnotesize{(3.09)} & 
        37.84 \footnotesize{(1.39)} \\
        {\scshape Subset$_{\text{Ent}}$} & 
        \underline{34.28} \footnotesize{(1.52)} & 
        34.62 \footnotesize{(2.56)} & 
        42.00 \footnotesize{(1.71)} & 
        32.04 \footnotesize{(2.73)} & 
        31.03 \footnotesize{(2.02)} & 
        38.00 \footnotesize{(1.33)} \\
        {\scshape Full-set} & 
        29.59 \footnotesize{(0.85)} & 
        36.92 \footnotesize{(1.43)} & 
        31.16 \footnotesize{(4.97)} & 
        33.72 \footnotesize{(8.36)} & 
        29.41 \footnotesize{(8.25)} & 
        34.37 \footnotesize{(8.93)} \\ 
        \midrule
        {\scshape Apa$_{\text{Fixed}}$} & 
        31.74 \footnotesize{(1.16)} & 
        \underline{39.63} \footnotesize{(0.89)} & 
        \textbf{45.01} \footnotesize{(2.06)} & 
        \textbf{43.95} \footnotesize{(2.07)} & 
        \textbf{36.45} \footnotesize{(0.38)} & 
        \textbf{42.18} \footnotesize{(3.37)} \\
        {\scshape Apa$_{\text{Gen}}$} & 
        \textbf{34.38} \footnotesize{(0.40)} & 
        \textbf{41.89} \footnotesize{(2.02)} & 
        \underline{43.29} \footnotesize{(3.69)} & 
        \underline{40.70} \footnotesize{(2.98)} & 
        \underline{31.72} \footnotesize{(3.24)} & 
        \underline{39.36} \footnotesize{(1.45)} \\

        
        \midrule
        \multicolumn{7}{l}{\textbf{AmbigTriviaQA}} \\
        \midrule
        {\scshape Subset$_{\text{Rand}}$} & 
        54.71 \footnotesize{(2.26)} & 
        70.97 \footnotesize{(2.57)} & 
        60.57 \footnotesize{(0.81)} & 
        67.82 \footnotesize{(4.14)} & 
        63.19 \footnotesize{(3.06)} & 
        73.52 \footnotesize{(3.94)} \\
        {\scshape Subset$_{\text{Ent}}$} & 
        58.83 \footnotesize{(1.42)} & 
        74.98 \footnotesize{(2.09)} &
        62.17 \footnotesize{(0.81)} & 
        67.16 \footnotesize{(4.14)} & 
        64.95 \footnotesize{(1.17)} & 
        76.03 \footnotesize{(0.86)} \\
        {\scshape Full-set} & 
        58.10 \footnotesize{(0.66)} & 
        71.25 \footnotesize{(1.53)} & 
        66.67 \footnotesize{(0.66)} & 
        76.38 \footnotesize{(0.53)} & 
        68.33 \footnotesize{(0.82)} & 
        76.82 \footnotesize{(0.91)} \\ 
        \midrule
        {\scshape Apa$_{\text{Fixed}}$} & 
        \textbf{62.97} \footnotesize{(0.63)} & 
        \underline{75.50} \footnotesize{(0.62)} & 
        \textbf{70.70} \footnotesize{(1.16)} & 
        \textbf{83.48} \footnotesize{(0.59)} & 
        \textbf{70.83} \footnotesize{(1.43)} & 
        \textbf{80.99} \footnotesize{(1.67)} \\
        {\scshape Apa$_{\text{Gen}}$} & 
        \underline{59.27} \footnotesize{(1.07)} & 
        \textbf{75.74} \footnotesize{(1.52)} & 
        \underline{67.73} \footnotesize{(1.11)} & 
        \underline{82.14} \footnotesize{(1.76)} & 
        \underline{69.25} \footnotesize{(1.59)} & 
        \underline{79.57} \footnotesize{(1.74)} \\

        
        \midrule
        \multicolumn{7}{l}{\textbf{AmbigWebQuestions}} \\
        \midrule
        {\scshape Subset$_{\text{Rand}}$} & 
        38.69 \footnotesize{(1.83)} & 
        73.84 \footnotesize{(1.67)} & 
        45.16 \footnotesize{(2.03)} & 
        71.74 \footnotesize{(1.75)} & 
        44.31 \footnotesize{(3.51)} & 
        72.99 \footnotesize{(2.36)} \\
        {\scshape Subset$_{\text{Ent}}$} & 
        42.39 \footnotesize{(1.36)} & 
        75.86 \footnotesize{(0.94)} & 
        50.93 \footnotesize{(5.43)} & 
        71.11 \footnotesize{(4.74)} & 
        48.70 \footnotesize{(1.19)} & 
        77.43 \footnotesize{(1.34)} \\
        {\scshape Full-set} &
        40.46 \footnotesize{(4.04)} & 
        73.84 \footnotesize{(1.67)} & 
        41.83 \footnotesize{(1.95)} & 
        74.72 \footnotesize{(0.40)} & 
        47.20 \footnotesize{(1.59)} & 
        75.27 \footnotesize{(0.75)} \\ 
        \midrule
        {\scshape Apa$_{\text{Fixed}}$} & 
        \textbf{49.15} \footnotesize{(2.57)} & 
        \textbf{77.07} \footnotesize{(1.67)} & 
        \textbf{54.02} \footnotesize{(2.17)} & 
        \textbf{81.07} \footnotesize{(1.26)} & 
        \textbf{53.69} \footnotesize{(0.97)} & 
        \textbf{79.22} \footnotesize{(0.35)} \\
        {\scshape Apa$_{\text{Gen}}$} & 
        \underline{47.26} \footnotesize{(1.01)} & 
        \underline{76.64} \footnotesize{(0.50)} & 
        \underline{51.41} \footnotesize{(0.92)} & 
        \underline{79.54} \footnotesize{(0.24)} & 
        \underline{52.96} \footnotesize{(3.46)} & 
        \underline{78.46} \footnotesize{(2.00)} \\

        
         \midrule
        \multicolumn{7}{l}{\textbf{AmbigFreebaseQA}} \\
        \midrule
        {\scshape Subset$_{\text{Rand}}$} & 
        63.59 \footnotesize{(2.53)} & 
        77.70 \footnotesize{(1.93)} & 
        70.60 \footnotesize{(1.27)} & 
        75.93 \footnotesize{(4.66)} & 
        70.40 \footnotesize{(7.06)} & 
        78.29 \footnotesize{(5.35)} \\
        {\scshape Subset$_{\text{Ent}}$} & 
        72.18 \footnotesize{(0.87)} & 
        83.89 \footnotesize{(1.10)} & 
        72.94 \footnotesize{(2.97)} & 
        77.17 \footnotesize{(4.66)} & 
        73.38 \footnotesize{(0.89)} & 
        81.93 \footnotesize{(0.25)} \\
        {\scshape Full-set} & 
        69.97 \footnotesize{(1.33)} & 
        80.34 \footnotesize{(1.19)} & 
        76.98 \footnotesize{(2.62)} & 
        84.67 \footnotesize{(3.08)} & 
        76.56 \footnotesize{(1.13)} & 
        83.00 \footnotesize{(0.69)} \\ 
        \midrule
        {\scshape Apa$_{\text{Fixed}}$} & 
        \textbf{73.37} \footnotesize{(0.40)} & 
        \underline{84.19} \footnotesize{(0.45)} & 
        \textbf{80.84} \footnotesize{(0.69)} & 
        \textbf{90.12} \footnotesize{(0.27)} & 
        \textbf{79.92} \footnotesize{(2.82)} & 
        \textbf{88.03} \footnotesize{(1.51)} \\
        {\scshape Apa$_{\text{Gen}}$} & 
        \underline{73.18} \footnotesize{(0.74)} & 
        \textbf{84.90} \footnotesize{(0.40)} & 
        \underline{80.27} \footnotesize{(1.32)} & 
        \underline{89.22} \footnotesize{(0.96)} & 
        \underline{79.80} \footnotesize{(2.14)} & 
        \underline{87.61} \footnotesize{(2.82)}\\ 
        \bottomrule
        
        \end{tabular}

    }
    \caption{
        Average and standard deviation (in parentheses) of the trained methods over three different random seeds.
        The \textbf{best method} is highlighted in bold and the \underline{second-best method} is underlined.
    }
    \label{table:full_results}
\end{table*}

\end{document}